\documentclass[dvipsnames]{article} 
\usepackage{iclr2021_conference,times}

\usepackage{amsmath,amsfonts,bm}

\def\eqref#1{equation~\ref{#1}}

\def\1{\bm{1}}

\def\vs{{\bm{s}}}

\DeclareMathAlphabet{\mathsfit}{\encodingdefault}{\sfdefault}{m}{sl}
\SetMathAlphabet{\mathsfit}{bold}{\encodingdefault}{\sfdefault}{bx}{n}

\usepackage{hyperref}
\usepackage{url}

\usepackage{comment}
\usepackage{amsmath,amssymb} 
\usepackage{subcaption}

\usepackage{graphicx}
\usepackage[font=footnotesize,labelfont=bf]{caption}
\usepackage{amsmath,booktabs}

\usepackage{soul}
\usepackage{hyperref}
\usepackage{enumitem}
\usepackage{ifthen} 
\usepackage{bbm}

\newcommand{\ie}{\textit{i.e.~}}

\renewcommand{\vs}{\textit{vs.~}}
\newcommand{\na}{ {\tiny \color{gray}{\textit{n/a}} } }

\usepackage{pifont}
\newcommand{\cmark}{\ding{51}}%
\newcommand{\xmark}{\color{lightgray}{\ding{55}}}%

\title{NBDT: Neural-Backed Decision Tree}

\author{Alvin Wan$_1$, Lisa Dunlap\textbf{\thanks{denotes equal contribution}}$_1$~, Daniel Ho$^*_1$, Jihan Yin$_1$, Scott Lee$_1$, Suzanne Petryk$_1$, \\ \textbf{Sarah Adel Bargal$_2$, Joseph E. Gonzalez$_1$} \\
UC Berkeley$_1$, Boston University$_2$\\
{\scriptsize \texttt{\{alvinwan,ldunlap,danielho,jihan\_yin,scott.lee.3898,spetryk,jegonzal\}@berkeley.edu}}\\
{\scriptsize \texttt{sbargal@bu.edu}}
}

\iclrfinalcopy 
\begin{document}

\maketitle

\begin{abstract}
Machine learning applications such as finance and medicine demand accurate and justifiable predictions, barring most deep learning methods from use. In response, previous work combines decision trees with deep learning, yielding models that (1) sacrifice interpretability for accuracy or (2) sacrifice accuracy for interpretability. We forgo this dilemma by \textit{jointly improving} accuracy and interpretability using Neural-Backed Decision Trees (NBDTs).
NBDTs replace a neural network's final linear layer with a differentiable sequence of decisions and a surrogate loss.
This forces the model to learn high-level concepts and lessens reliance on highly-uncertain decisions,
 yielding (1) accuracy: NBDTs match or outperform modern neural networks on CIFAR, ImageNet and better generalize to unseen classes by up to 16\%. Furthermore, our surrogate loss improves the \textit{original} model's accuracy by up to 2\%.
NBDTs also afford (2) interpretability: improving human trust by clearly identifying model mistakes and assisting in dataset debugging. Code and pretrained NBDTs are at \href{https://github.com/alvinwan/neural-backed-decision-trees}{\color{blue}{github.com/alvinwan/neural-backed-decision-trees}}.

\end{abstract}

\section{Introduction}

Many computer vision applications (e.g. medical imaging and autonomous driving) require insight into the model's decision process, complicating applications of deep learning which are traditionally black box.
Recent efforts in explainable computer vision attempt to address this need
and can be grouped into one of two categories: (1) saliency maps and (2) sequential decision processes. Saliency maps retroactively explain model predictions by identifying which pixels most affected the prediction.
However, by focusing on the input, saliency maps fail to capture the model's decision making process.
For example, saliency offers no insight for a misclassification when the model is ``looking'' at the right object for the wrong reasons. 
Alternatively, we can gain insight into the model's decision process by breaking up predictions into a sequence of smaller semantically meaningful decisions as in rule-based models like decision trees. 
However, existing efforts to fuse deep learning and decision trees suffer from 
(1) significant accuracy loss, 
relative to contemporary models (e.g., residual networks),
(2) reduced interpretability due to accuracy optimizations (e.g., impure leaves and ensembles), and
(3) tree structures that offer limited insight into the model's credibility.

To address these, we propose \textbf{Neural-Backed Decision Trees (NBDTs)} to jointly improve \textit{both} (1) accuracy and (2) interpretability of modern neural networks, utilizing decision rules that preserve (3) properties like sequential, discrete decisions; pure leaves; and non-ensembled predictions. These properties in unison enable unique insights, as we show.
We acknowledge that there is no universally-accepted definition of interpretability \citep{lundberg2020local,doshi2017towards,lipton2016mythos}, so to show interpretability, we adopt a definition offered by \cite{measuringinterp}: A model is interpretable if a human can validate its prediction, determining when the model has made a sizable mistake. We picked this definition for its importance to downstream benefits we can evaluate, specifically (1) model or dataset debugging and (2) improving human trust.
To accomplish this, NBDTs replace the final linear layer of a neural network with a differentiable oblique decision tree and, unlike its predecessors (\ie decision trees, hierarchical classifiers), uses a hierarchy derived from model parameters, does not employ a hierarchical softmax, and can be created from \textit{any} existing classification neural network without architectural modifications. These improvements tailor the hierarchy to the network rather than overfit to the feature space, lessens the decision tree's reliance on highly uncertain decisions, and encourages accurate recognition of high-level concepts. These benefits culminate in joint improvement of accuracy and interpretability. Our contributions:

\begin{enumerate}
    \item We propose a \textit{tree supervision loss}, yielding NBDTs that match/outperform and out-generalize modern neural networks (WideResNet, EfficientNet) on ImageNet, TinyImageNet200, and CIFAR100. Our loss also improves the \textit{original} model by up to 2\%.
    \item We propose alternative hierarchies for oblique decision trees -- \textit{induced hierarchies} built using pre-trained neural network weights -- that outperform both data-based hierarchies (e.g. built with information gain) and existing hierarchies (e.g. WordNet), in accuracy.
    \item We show NBDT explanations are more helpful to the user when identifying model mistakes, preferred when using the model to assist in challenging classification tasks, and can be used to identify ambiguous ImageNet labels.  
\end{enumerate}

\section{Related Works}

\textbf{Saliency Maps.} 
Numerous efforts~\citep{grad,deconv,simonyan2013deep,zhang2016top,gradcam,lime,rise,ig} have explored the design of saliency maps identifying pixels that most influenced \textbf{}the model's prediction. 
White-box techniques \citep{grad,deconv,simonyan2013deep,gradcam,ig} use the network's parameters to determine salient image regions, and black-box techniques \citep{lime,rise} determine pixel importance by measuring the prediction's response to perturbed inputs. However, saliency does not explain the model's decision process (e.g. Was the model confused early on, distinguishing between \textit{Animal} and \textit{Vehicle}? Or is it only confused between dog breeds?). 

\textbf{Transfer to Explainable Models.} Prior to the recent success of deep learning, decision trees were state-of-the-art on a wide variety of learning tasks and the gold standard for interpretability. Despite this recency, study at the intersection of neural network and decision tree dates back three decades, where neural networks were seeded with decision tree weights \citep{1990_dt_init,1994_dt_init,1995_dt_init,1995_dt_nn}, and decision trees were created from neural network queries \citep{1999_extract_dt,dectext,decision_tree_extraction,craven1996extracting,craven1994using}, like distillation \citep{hinton2015distilling}. The modern analog of both sets of work \citep{2018_dt_init,tree_grad,soft_decision_tree} evaluate on feature-sparse, sample-sparse regimes such as the UCI datasets~\citep{uci} or MNIST~\citep{lecun2010mnist} and \textit{perform poorly} on standard image classification tasks.

\textbf{Hybrid Models.} Recent work produces hybrid decision tree and neural network models to scale up to datasets like CIFAR10~\citep{cifar},  CIFAR100~\citep{cifar}, TinyImageNet~\citep{Le2015TinyIV}, and ImageNet~\citep{imagenet}. One category of models organizes the neural network into a hierarchy, dynamically selecting branches to run inference \citep{adaptive_inference_graphs,dynamic_routing,hydranets,yolo9000,blockout}. However, these models use \textit{impure leaves} resulting in uninterpretatble, stochastic paths. Other approaches fuse deep learning into each decision tree node: an entire neural network \citep{deep_decision_network}, several layers \citep{blockout,neural_regression_forest}, a linear layer \citep{network_of_experts}, or some other parameterization of neural network output \citep{deep_neural_decision_forests}. These models see reduced interpretability by using k-way decisions with large k (via depth-2 trees) \citep{network_of_experts,cnnrnn} or employing an ensemble~\citep{deep_neural_decision_forests,network_of_experts}, which is often referred to as a ``black box'' \citep{carvalho2019machine,rudin2018stop}.

\textbf{Hierarchical Classification} \citep{silla2011survey}. One set of approaches directly uses a pre-existing hierarchy over classes, such as WordNet \citep{yolo9000,brust2019integrating,denglarge}. However \textit{conceptual similarity is not indicative of visual similarity}. Other models build a hierarchy using the training set directly, via a classic data-dependent metric like Gini impurity \citep{xoc} or information gain \citep{neural_decision_forest,cign}. These models are instead \textit{prone to overfitting}, per \citep{adaptive_neural_trees}. Finally, several works introduce hierarchical surrogate losses \citep{treeregularization,hedging}, such as hierarchical softmax \citep{mohammed2018effectiveness}, but as the authors note, these methods quickly suffer from major accuracy loss with more classes or higher-resolution images (e.g. beyond CIFAR10). We demonstrate hierarchical classifiers attain higher accuracy \textit{without} a hierarchical softmax.

\section{Method} \label{sec:method}

Neural-Backed Decision Trees (NBDTs) replace a network's final linear layer with a decision tree. Unlike classical decision trees or many hierarchical classifiers, NBDTs use path probabilities for inference (Sec \ref{sec:method-prediction}) to tolerate highly-uncertain intermediate decisions, build a hierarchy from pre-trained model weights (Sec \ref{sec:method-building} \& \ref{sec:labeling-decision-nodes}) to lessen overfitting, and train with a hierarchical loss (Sec \ref{sec:method-supervision}) to significantly better learn high-level decisions (e.g., \textit{Animal} vs. \textit{Vehicle}).

\subsection{Inference}
\label{sec:method-prediction}

Our NBDT first featurizes each sample using the neural network backbone; the backbone consists of all neural network layers before the final linear layer. Second, we run the final fully-connected layer as an oblique decision tree. However, (a) a classic decision tree cannot recover from a mistake early in the hierarchy and (b) just running a classic decision tree on neural features drops accuracy significantly, by up to 11\% (Table \ref{tab:hierarchies}). Thus, we present modified decision rules (Figure \ref{fig:inference-modes}, B):

\begin{figure}[t]
    \centering
  \includegraphics[width=\linewidth]{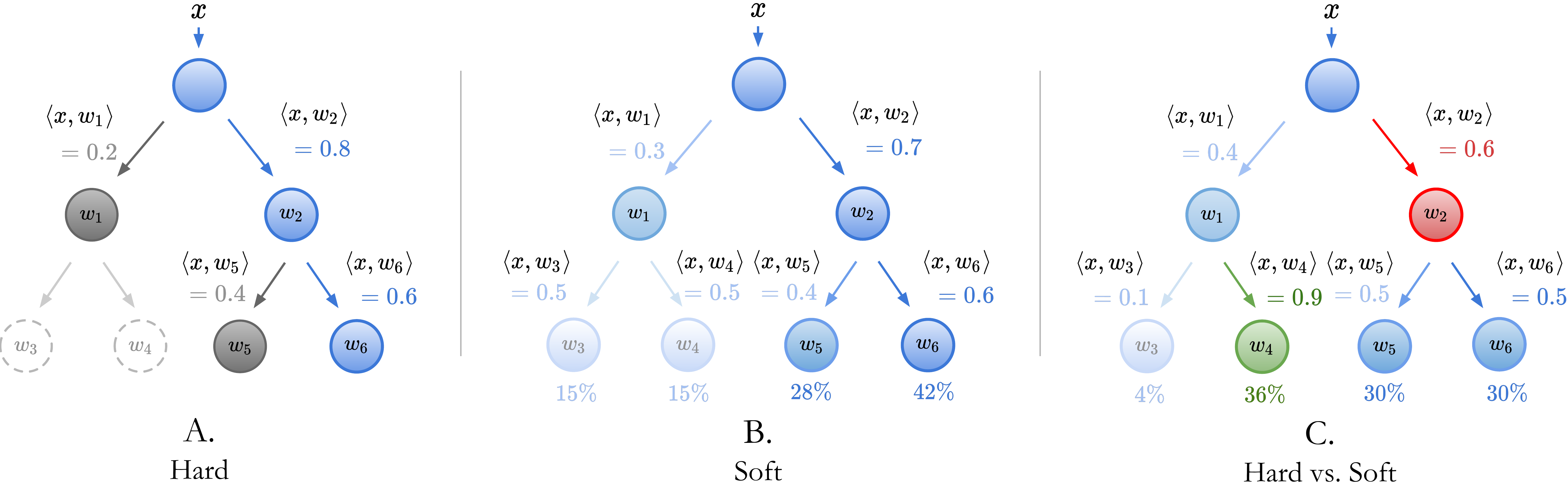}
  \caption{\textbf{Hard and Soft Decision Trees.} \textbf{A. Hard:} is the classic ``hard'' oblique decision tree. Each node picks the child node with the largest inner product, and visits that node next. Continue until a leaf is reached. \textbf{B. Soft:} is the ``soft'' variant, where each node simply returns probabilities, as normalized inner products, of each child. For each leaf, compute the probability of its path to the root. Pick leaf with the highest probability. \textbf{C. Hard vs. Soft:} Assume $w_4$ is the correct class. With hard inference,  the mistake at the root (red) is irrecoverable. However, with soft inference, the highly-uncertain decisions at the root and at $w_2$ are superseded by the highly certain decision at $w_3$ (green). This means the model can still correctly pick $w_4$ despite a mistake at the root. In short, soft inference can tolerate mistakes in highly uncertain decisions.
  }
  \label{fig:inference-modes}
\end{figure}

\textbf{1. Seed oblique decision rule weights with neural network weights.} An oblique decision tree supports only binary decisions, using a hyperplane for each decision. Instead, we associate a weight vector $n_i$ with each node. For leaf nodes, where $i=k \in [1, K]$, each $n_i = w_k$ is a row vector from the fully-connected layer's weights $W\in\mathbb{R}^{D\times K}$. For all inner nodes, where $i \in [K+1, N]$, find all leaves $k \in L(i)$ in node $i$'s subtree and average their weights: $n_i = \sum_{k \in L(i)} w_k / |L(i)|$.

\textbf{2. Compute node probabilities.} Child probabilities are given by softmax inner products. For each sample $x$ and node $i$, compute the probability of each child $j \in C(i)$ using $p(j|i) = \textsc{Softmax}(\langle \vec{n}_i, x\rangle)[j]$, where $\vec{n}_i = (\langle n_j, x\rangle)_{j \in C(i)}$.

\textbf{3. Pick a leaf using path probabilities.} Inspired by \cite{hedging}, consider a leaf, its class $k$ and its path from the root $P_k$.
The probability of each node $i \in P_k$ traversing the next node in the path $C_k(i) \in P_k \cap C(i)$ is denoted $p(C_k(i)|i)$. Then, the probability of leaf and its class $k$ is

\begin{equation}\label{eqn:soft_prob_leaf}
    p(k) = \Pi_{i \in P_k} p(C_k(i)|i)
\end{equation}

In soft inference, the final class prediction $\hat{k}$ is defined over these class probabilities, 

\begin{equation}\label{eqn:soft_inference}
    \hat{k} = \text{argmax}_k p(k) = \text{argmax}_k \Pi_{i \in P_k} p(C_k(i)|i)
\end{equation}

Our inference strategy has two benefits: (a) Since the architecture is unchanged, the fully-connected layer can be run regularly (Table \ref{tab:onn}) or as decision rules (Table \ref{tab:cifar}), and (b) unlike decision trees and other conditionally-executed models \citep{adaptive_neural_trees,adaptive_inference_graphs}, our method can recover from a mistake early in the hierarchy with sufficient uncertainty in the incorrect path (Figure \ref{fig:inference-modes} C, Appendix Table \ref{tab:hard-inference}).
This inference mode bests classic tree inference (Appendix \ref{sec:hard-inference}).

\subsection{Building Induced Hierarchies}
\label{sec:method-building}

Existing decision-tree-based methods use (a) hierarchies built with data-dependent heuristics like information gain or (b) existing hierarchies like WordNet. However, the former overfits to the data, and the latter focuses on conceptual rather than visual similarity: For example, by virtue of being an animal, \textit{Bird} is closer to \textit{Cat} than to \textit{Plane}, according to WordNet. However, the opposite is true for visual similarity: by virtue of being in the sky, \textit{Bird} is more visually similar to \textit{Plane} than to \textit{Cat}. Thus, to prevent overfitting and reflect visual similarity, we build a hierarchy using model weights.

\begin{figure}[t]
    \centering
  \includegraphics[width=\linewidth]{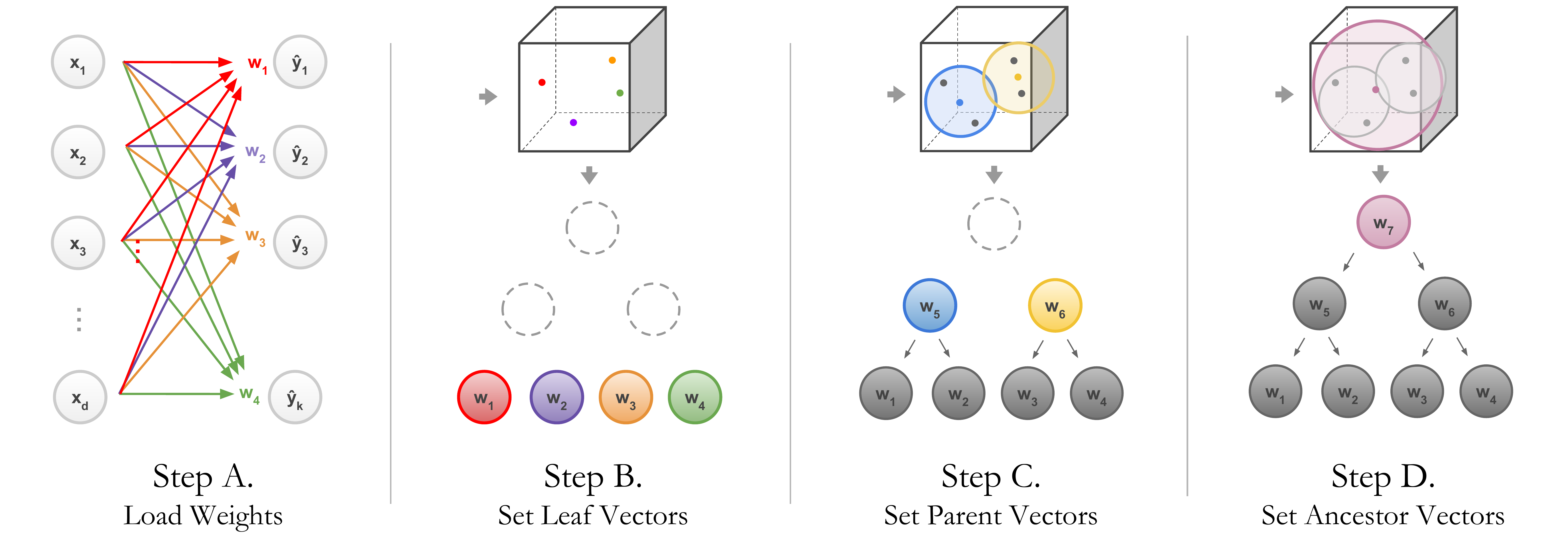}
  \caption{\textbf{Building Induced Hierarchies.} \textbf{Step A.} Load the weights of a pre-trained model's final fully-connected layer, with weight matrix $W \in \mathbb{R}^{D \times K}$. \textbf{Step B.} Take rows $w_k \in W$ and normalize for each leaf node's weight. For example, the red $w_1$ in A is assigned to the red leaf in B. \textbf{Step C.} Average each pair of leaf nodes for the parents' weight. For example, $w_1$ and $w_2$ (red and purple) in B are averaged to make $w_5$ (blue) in C. \textbf{Step D.} For each ancestor, average all leaf node weights in its subtree. That average is the ancestor's weight. Here, the ancestor \textit{is} the root, so its weight is the average of all leaf weights $w_1, w_2, w_3, w_4$.
  }
  \label{fig:induced_tree_fig}
\end{figure}

Our hierarchy requires pre-trained model weights. Take row vectors $w_k: k \in [1, K]$, each representing a class, from the fully-connected layer weights $W$. Then, run hierarchical agglomerative clustering on the normalized class representatives $w_k / \|w_k\|_2$. Agglomerative clustering decides which nodes and groups of nodes are iteratively paired. As described in Sec \ref{sec:method-prediction}, each leaf node's weight is a row vector $w_k \in W$ (Figure \ref{fig:induced_tree_fig}, Step B) and each inner node's weight $n_i$ is the average of its leaf node's weights (Figure \ref{fig:induced_tree_fig}, Step C). This hierarchy is the \textit{induced hierarchy} (Figure \ref{fig:induced_tree_fig}).

\subsection{Labeling Decision Nodes with WordNet}
\label{sec:labeling-decision-nodes}

WordNet is a hierarchy of nouns. To assign WordNet meaning to nodes, we compute the earliest common ancestor for all leaves in a subtree: For example, say \textit{Dog} and \textit{Cat} are two leaves that share a parent. To find WordNet meaning for the parent, find all ancestor concepts that \textit{Dog} and \textit{Cat} share, like \textit{Mammal}, \textit{Animal}, and \textit{Living Thing}. The earliest shared ancestor is \textit{Mammal}, so we assign \textit{Mammal} to the parent of \textit{Dog} and \textit{Cat}. We repeat for all inner nodes.

However, the WordNet corpus is lacking in concepts that are not themselves objects, like object attributes (e.g., \textit{Pencil} and \textit{Wire} are both cylindrical) and (b) abstract visual ideas like context (e.g., \textit{fish} and \textit{boat} are both aquatic). Many of these which are littered across our induced hierarchies (Appendix Figure \ref{fig:cifar100_tree}). Despite this limitation, we use WordNet to assign meaning to intermediate decision nodes, with more sophisticated methods left to future work.

\subsection{Fine-tuning with Tree Supervision Loss}
\label{sec:method-supervision}

Even though standard cross entropy loss separates representatives for each leaf, it is not trained to separate representatives for each inner node (Table \ref{tab:losses}, ``None''). To amend this, we add a \textit{tree supervision loss}, a cross entropy loss over the class distribution of path probabilities $\mathcal{D}_\text{nbdt} = \{p(k)\}_{k = 1}^K$ (Eq. \ref{eqn:soft_prob_leaf}) from Sec \ref{sec:method-prediction}, with time-varying weights $\omega_t, \beta_t$ where $t$ is the epoch count:

\begin{equation}
    \mathcal{L} = \beta_t\underbrace{\textsc{CrossEntropy}(\mathcal{D}_\text{pred}, \mathcal{D}_\text{label})}_{\mathcal{L}_\text{original}} + 
    \omega_t\underbrace{\textsc{CrossEntropy}(\mathcal{D}_\text{nbdt}, \mathcal{D}_\text{label})}_{\mathcal{L}_\text{soft}}
\end{equation}

Our tree supervision loss $\mathcal{L}_\text{soft}$ requires a pre-defined hierarchy. We find that (a) tree supervision loss damages learning speed early in training, when leaf weights are nonsensical. Thus, our tree supervision weight $\omega_t$ grows linearly from $\omega_0=0$ to $\omega_T=0.5$ for CIFAR10, CIFAR100, and to $\omega_T=5$ for TinyImageNet, ImageNet; $\beta_t \in [0, 1]$ decays linearly over time.
(b) We re-train where possible, fine-tuning with $\mathcal{L}_\text{soft}$ only when the original model accuracy is not reproducible.
(c) Unlike hierarchical softmax, our path-probability cross entropy loss $\mathcal{L}_\text{soft}$ disproportionately up-weights decisions earlier in the hierarchy, encouraging accurate high-level decisions; this is reflected our out-generalization of the baseline neural network by up to 16\% to unseen classes (Table \ref{tab:generalization}).

\section{Experiments}

NBDTs obtain state-of-the-art results for interpretable models and match or outperform modern neural networks on image classification. We report results on different models (ResNet, WideResNet, EfficientNet) and datasets (CIFAR10, CIFAR100, TinyImageNet, ImageNet). We additionally conduct ablation studies to verify the hierarchy and loss designs, find that our training procedure improves the \textit{original} neural network's accuracy by up to 2\%, and show that NBDTs improve generalization to unseen classes by up to 16\%. All reported improvements are absolute.

\subsection{Results}

\textbf{Small-scale Datasets.} Our method (Table \ref{tab:cifar}) matches or outperforms recently state-of-the-art neural networks. On CIFAR10 and TinyImageNet, NBDT accuracy falls within 0.15\% of the baseline neural network. On CIFAR100, NBDT accuracy outperforms the baseline by $\sim$1\%.

\textbf{Large-scale Dataset.} On ImageNet (Table \ref{tab:imagenet}), NBDTs obtain 76.60\% top-1 accuracy, outperforming the strongest competitor NofE by ~15\%. Note that we take the best competing results for any decision-tree-based method, but the strongest competitors hinder interpretability by using ensembles of models like a decision forest (DNDF, DCDJ) or feature shallow trees with only depth 2 (NofE).

\begin{table}[t]
\small
\centering
\caption{\textbf{Results.} NBDT outperforms competing decision-tree-based methods by up to 18\% and can also outperform the \textit{original} neural network by $\sim1$\%. ``Expl?'' indicates the method retains interpretable properties: pure leaves, sequential decisions, non-ensemble. Methods without this check see reduced interpretability. We bold the highest decision-tree-based accuracy. These results are taken directly from the original papers (\textit{n/a} denotes results missing from original papers): XOC \citep{xoc}, DCDJ \citep{decision_jungle}, NofE \citep{network_of_experts}, DDN \citep{deep_decision_network}, ANT \citep{adaptive_neural_trees}, CNN-RNN \citep{cnnrnn}. We train DNDF \citep{deep_neural_decision_forests} with an updated R18 backbone, as they did not report CIFAR accuracy.
}

\vspace{5pt}
\begin{tabular*}{\textwidth}{l @{\extracolsep{\fill}} llllll}
\toprule
Method & Backbone     & Expl? & CIFAR10 & CIFAR100 & TinyImageNet \\
\midrule
NN & WideResNet28x10 & \xmark & \textit{97.62\%} & \textit{82.09\%}  & \textit{67.65\%} \\
ANT-A* & \na & \cmark & 93.28\% & \na  & \na \\

DDN & NiN & \xmark & 90.32\% & 68.35\%  & \na \\
DCDJ & NiN & \xmark & \na & 69.0\%  & \na \\
NofE & ResNet56-4x & \xmark & \na & 76.24\%  & \na \\

CNN-RNN & WideResNet28x10 & \cmark & \na & 76.23\% & \na \\
NBDT-S (Ours) &  WideResNet28x10 & \cmark & \textbf{97.55\%} & \textbf{82.97\%}  & \textbf{67.72\%}\\

\midrule
NN & ResNet18 & \xmark & \textit{94.97\%} & \textit{75.92\%}  & \textit{64.13\%}\\
DNDF & ResNet18  & \xmark & 94.32\% & 67.18\% & 44.56\%\\
XOC &  ResNet18 & \cmark & 93.12\% & \na  & \na \\
DT & ResNet18 & \cmark & 93.97\% & 64.45\% & 52.09\% \\
NBDT-S (Ours) & ResNet18     & \cmark & \textbf{94.82}\% & \textbf{77.09\%} & \textbf{64.23\%}  \\

\bottomrule
\end{tabular*}
\label{tab:cifar}
\end{table}

\begin{figure*}[t]
    \centering
    \vspace{5pt}
    \small
    \caption{\textbf{ImageNet Results.} NBDT outperforms all competing decision-tree-based methods by at least 14\%, staying within 0.6\% of EfficientNet accuracy. ``EfficientNet'' is EfficientNet-EdgeTPU-Small.}
        
    \vspace{5pt}
    \begin{tabular*}{\textwidth}{l @{\extracolsep{\fill}} llllll}
        \toprule
        Method & NBDT (ours) & NBDT (ours) & XOC & NofE \\
        \midrule
        Backbone & EfficientNet & ResNet18 & ResNet152 & AlexNet \\
        Original Acc & 77.23\% & 60.76\% & 78.31\% & 56.55\% \\
        Delta Acc & -0.63\% & +0.50\% & -17.5\% & +4.7\% \\
        Explainable Acc & \textbf{76.60\%} & 61.26\% & 60.77\% & 61.29\%\\
        \bottomrule
    \end{tabular*}
    \label{tab:imagenet}
\end{figure*}

\subsection{Analysis}
\label{sec:analysis}

Analyses show that our NBDT improvements are dominated by significantly improved ability to distinguish higher-level concepts (e.g., \textit{Animal} vs. \textit{Vehicle}).

\textbf{Comparison of Hierarchies.} Table \ref{tab:hierarchies} shows that our induced hierarchies outperform alternatives. In particular, \textit{data-dependent} hierarchies overfit, and the existing \textit{WordNet hierarchy} focuses on conceptual rather than visual similarity.

\textbf{Comparisons of Losses.} Previous work suggests hierarchical softmax (Appendix \ref{sec:hard-tree-supervision-loss}) is necessary for hierarchical classifiers. However, our results suggest otherwise: NBDTs trained with hierarchical softmax see $\sim$3\% less accuracy than with tree supervision loss on TinyImageNet (Table \ref{tab:losses}).

\begin{table}[t]
\small
\centering
\caption{\textbf{Comparisons of Hierarchies.} We demonstrate that our weight-space hierarchy bests taxonomy and data-dependent hierarchies. In particular, the induced hierarchy achieves better performance than (a) the WordNet hierarchy, (b) a classic decision tree's information gain hierarchy, built over neural features (``Info Gain''), and (c) an oblique decision tree built over neural features (``OC1'').
}
\vspace{5pt}
\begin{tabular*}{\textwidth}{l @{\extracolsep{\fill}} llllll}
\toprule
Dataset & Backbone     & Original & Induced & Info Gain & WordNet & OC1
\\
\midrule
CIFAR10 & ResNet18 & \textit{94.97\%} & \textbf{94.82\%} & 93.97\% & 94.37\% & 94.33\% \\
CIFAR100 & ResNet18 & \textit{75.92\%} & \textbf{77.09\%} & 64.45\% & 74.08\% & 38.67\% \\
TinyImageNet200 & ResNet18 & \textit{64.13\%} & \textbf{64.23\%} & 52.09\% & 60.26\% & 15.63\% \\
\bottomrule
\end{tabular*}
\label{tab:hierarchies}
\end{table}

\begin{table}[t]
\small
\centering
\caption{\textbf{Comparisons of Losses.} Training the NBDT using tree supervision loss with a linearly increasing weight (``TreeSup(t)'') is superior to training (a) with a constant-weight tree supervision loss (``TreeSup''), (b) with a hierarchical softmax (``HrchSmax'') and (c) without extra loss terms. (``None''). $\Delta$ is the accuracy difference between our soft loss and hierarchical softmax.}
\vspace{5pt}
\begin{tabular*}{\textwidth}{l @{\extracolsep{\fill}} lllllll}
\toprule
Dataset & Backbone     &  Original & TreeSup(t) & TreeSup & None & HrchSmax \\
\midrule
CIFAR10 & ResNet18 & \textit{94.97\%} & \textbf{94.82\%}& 94.76\% & 94.38\% & 93.97\% \\
CIFAR100 & ResNet18 & \textit{75.92\%} & \textbf{77.09\%} & 74.92\% & 61.93\% & 74.09\% \\
TinyImageNet200 & ResNet18 & \textit{64.13\%} & \textbf{64.23\%} & 62.74\% & 45.51\% & 61.12\% \\
\bottomrule
\end{tabular*}
\label{tab:losses}
\end{table}

\begin{table}[t]
\small
\centering
\caption{\textbf{Mid-Training Hierarchy.} Constructing and using hierarchies early and often in training yields the highest performing models. All experiments use ResNet18 backbones. Per Sec \ref{sec:method-supervision}, $\beta_t, \omega_t$ are the loss term coefficients. Hierarchies are reconstructed every ``Period'' epochs, starting at ``Start'' and ending at ``End''.}
\vspace{5pt}
\begin{tabular*}{\textwidth}{lll@{\extracolsep{\fill}}lll@{\extracolsep{\fill}}lll}
\toprule
\multicolumn{3}{c}{Hierarchy Updates} & \multicolumn{3}{c}{CIFAR10} & \multicolumn{3}{c}{CIFAR100} \\
\cmidrule{1-3} \cmidrule{4-6} \cmidrule{7-9}
Start & End & Period & NBDT & NN+TSL & NN & NBDT & NN+TSL & NN \\
\midrule
67 & 120 & 10 & \textbf{94.88\%} & \textbf{94.97\%} & 94.97\% & \textbf{76.04\%} & \textbf{76.56\%} & 75.92\% \\
90 & 140 & 10 & 94.29\% & 94.84\%& 94.97\% & 75.44\% & 76.29\% & 75.92\% \\
90 & 140 & 20 & 94.52\% & 94.89\%& 94.97\% & 75.08\% & 76.11\% & 75.92\% \\
120 & 121 & 10 & 94.52\% & 94.92\%& 94.97\% & 74.97\% & 75.88\% & 75.92\% \\
\bottomrule
\end{tabular*}
\label{tab:mid-training-hierarchy}
\end{table}

\textbf{Original Neural Network.} Per Sec \ref{sec:method-prediction}, we can run the original neural network's fully-connected layer normally, after training with tree supervision loss. Using this, we find that the original neural network's accuracy improves by up to 2\% on CIFAR100, TinyImageNet (Table \ref{tab:onn}).

\textbf{Zero-Shot Superclass Generalization.} We define a ``superclass'' to be the hypernym of several classes. (e.g. \textit{Animal} is a superclass of \textit{Cat} and \textit{Dog}). Using WordNet (per Sec \ref{sec:method-building}), we (1) identify which superclasses each NBDT inner node is deciding between (e.g. \textit{Animal} \vs \textit{Vehicle}). (2) We find unseen classes that belong to the same superclass, from a different dataset. (e.g. Pull \textit{Turtle} images from ImageNet). (3) Evaluate the model to ensure the unseen class is classified into the correct superclass (e.g. ensure \textit{Turtle} is classified as \textit{Animal}). For an NBDT, this is straightforward: one of the inner nodes classifies \textit{Animal} \vs \textit{Vehicle} (Sec \ref{sec:labeling-decision-nodes}). For a standard neural network, we consider the superclass that the final prediction belongs to. (\ie When evaluating \textit{Animal} \vs \textit{Vehicle} on a \textit{Turtle} image, the CIFAR-trained model may predict any CIFAR \textit{Animal} class). See Appendix \ref{sec:explainability-node} for details. Our NBDT consistently bests the original neural network by 8\%+ (Table \ref{tab:generalization}). When discerning \textit{Carnivore vs. Ungulate}, NBDT outperforms the original neural network by 16\%.

\textbf{Mid-Training Hierarchy}:
We test NBDTs without using pre-trained weights, instead constructing hierarchies during training from the partially-trained network's weights. Tree supervision loss with mid-training hierarchies reliably improve the original neural network's accuracy, up to $\sim$0.6\%, and the NBDT itself can match the original neural network's accuracy (Table \ref{tab:mid-training-hierarchy}). However, this underperforms NBDT (Table \ref{tab:cifar}), showing fully-trained weights are still preferred for hierarchy construction.

\begin{table}
\parbox{.42\linewidth}{
    \centering
    \footnotesize
    \caption{\small\textbf{Original Neural Network.} We compare the model's accuracy before and after the tree supervision loss, using ResNet18, WideResNet on CIFAR100, TinyImageNet. Our loss increases the original network accuracy consistently by $\sim.8-2.4\%$. NN-S is the network trained with the tree supervision loss.}
    \begin{tabular*}{0.42\textwidth}{l @{\extracolsep{\fill}} llllll}
            \toprule
            Dataset & Backbone & NN & NN-S \\
            \midrule
            C100 & R18 & 75.92\% & \textbf{76.96\%} \\
            T200 & R18 & 64.13\% & \textbf{66.55\%} \\
            C100 & WRN28 & 82.09\% & \textbf{82.87\%} \\
            T200 & WRN28 & 67.65\% & \textbf{68.51\%} \\
            \bottomrule
        \end{tabular*}
     \label{tab:onn}
}
\hfill
\parbox{.53\linewidth}{
    \centering
    \footnotesize
    \caption{\small\textbf{Zero-Shot Superclass Generalization.} We evaluate a CIFAR10-trained NBDT (ResNet18 backbone) inner node's ability to generalize beyond seen classes. We label TinyImageNet with superclass labels (e.g. label \textit{Dog} with \textit{Animal}) and evaluate nodes distinguishing between said superclasses. We compare to the baseline ResNet18: check if the prediction is within the right superclass.}
    \begin{tabular*}{0.53\textwidth}{l @{\extracolsep{\fill}} llll}
            \toprule
            $n_\textrm{class}$     & Superclasses & R18 & NBDT-S \\ 
            \midrule
            71 & Animal \vs Vehicle & 66.08\% & \textbf{74.79\%} \\
            36 & Placental \vs Vertebrate & 45.50\% & \textbf{54.89\%} \\
            19 & Carnivore \vs Ungulate & 51.37\% & \textbf{67.78\%} \\ 
            9 & Motor Vehicle \vs Craft & 69.33\% & \textbf{77.78\%} \\
            \bottomrule
        \end{tabular*}
     \label{tab:generalization}
}
\end{table}

\section{Interpretability}

By breaking complex decisions into smaller intermediate decisions, decision trees provide insight into the decision process. 
However, when the intermediate decisions are themselves neural network predictions,
extracting insight becomes more challenging. 
To address this, we adopt benchmarks and an interpretability definition offered by \cite{measuringinterp}: A model is interpretable if a human can validate its prediction, determining when the model has made a sizable mistake. To assess this, we adapt \cite{measuringinterp}'s benchmarks to computer vision and show (a) humans can identify misclassifications with NBDT explanations more accurately than with saliency explanations (Sec \ref{sec:model_mistakes}), (b) a way to utilize NBDT's entropy to identify ambiguous labels (Sec. \ref{sec:dataset-debugging}), and (c) that humans prefer to agree with NBDT predictions when given a challenging image classification task (Sec. \ref{sec:blurry_task} \& \ref{sec:human-trust}).
Note that these analyses depend on three model properties that NBDT preserves: (1) discrete, sequential decisions, so that one path is selected; (2) pure leaves, so that one path picks one class; and (3) non-ensembled predictions, so that path to prediction attribution is discrete. In all surveys, we use CIFAR10-trained models with ResNet18 backbones.

\subsection{Survey: Identifying Faulty Model Predictions}
\label{sec:model_mistakes}
In this section we aim to answer a question posed in \citep{measuringinterp} \textit{"How well can someone detect when the model has made a sizable mistake?"}. In this survey, each user is given 3 images, 2 of which are correctly classified and 1 is mis-classified. Users must predict which image was incorrectly classified given a) the model explanations and b) \textit{without} the final prediction. For saliency maps, this is a near-impossible task as saliency usually highlights the main object in the image, regardless of wrong or right. However, hierarchical methods provide a sensible sequence of intermediate decisions that can be checked. This is reflected in the results: For each explainability technique, we collected \textbf{600} survey responses. When given saliency maps and class probabilities, only \textbf{87} predictions were correctly identified as wrong. In comparison, when given the NBDT series of predicted classes and child probabilities (e.g., “Animal (90\%) $\to$ Mammal (95\%)”, without the final leaf prediction) \textbf{237} images were correctly identified as wrong. Thus, respondents can better recognize mistakes in NBDT explanations nearly 3 times better.

Although NBDT provides more information than saliency maps about misclassification, a majority -- the remaining 363 NBDT predictions -- were not correctly identified. To explain this, we note that $\sim 37\%$ of all NBDT errors occur at the final binary decision, between two leaves; since we provide all decisions except the final one, these leaf errors would be impossible to distinguish.

\begin{figure}[t]
    \centering
    \begin{subfigure}{0.09\textwidth}
        \centering
        \includegraphics[width=\textwidth]{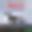}
    \end{subfigure}
    \begin{subfigure}{0.09\textwidth}
        \centering
        \includegraphics{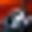}
    \end{subfigure}
    \begin{subfigure}{0.09\textwidth}
        \centering
        \includegraphics{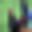}
    \end{subfigure}
    \begin{subfigure}{0.09\textwidth}
        \centering
        \includegraphics{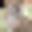}
    \end{subfigure}
    \begin{subfigure}{0.09\textwidth}
        \centering
        \includegraphics{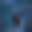}
    \end{subfigure}
    \begin{subfigure}{0.09\textwidth}
        \centering
        \includegraphics{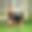}
    \end{subfigure}
    \begin{subfigure}{0.09\textwidth}
        \centering
        \includegraphics{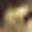}
    \end{subfigure}
    \begin{subfigure}{0.09\textwidth}
        \centering
        \includegraphics{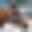}
    \end{subfigure}
    \begin{subfigure}{0.09\textwidth}
        \centering
        \includegraphics{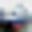}
    \end{subfigure}
    \begin{subfigure}{0.09\textwidth}
        \centering
        \includegraphics{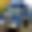}
    \end{subfigure}
    \caption{\textbf{CIFAR10 Blurry Images.} To make the classification task difficult for humans, the CIFAR10 images are downsampled by $4\times$. This forces at least partial reliance on model predictions, allowing us to evaluate which explanations are convincing enough to earn the user's agreement.}
    \label{fig:downsampled}
\end{figure}

\subsection{Survey: Explanation-Guided Image Classification}
\label{sec:blurry_task}
In this section we aim to answer a question posed in \citep{measuringinterp} \textit{“To what extent do people follow a model’s predictions when it is beneficial to do so?”}. In this first survey, each user is asked to classify a severely blurred image (Fig \ref{fig:downsampled}). This survey affirms the problem's difficulty, decimating human performance to not much more than guessing: \textbf{163} of \textbf{600} responses are correct (27.2\% accuracy).

In the next survey, we offer the blurred image and two sets of predictions: (1) the original neural network’s predicted class and its saliency map, and (2) the NBDT predicted class and the sequence of decisions that led up to it (“Animal, Mammal, Cat”). For all examples, the two models predict different classes. In 30\% of the examples, NBDT is right and the original model is wrong. In another 30\%, the opposite is true. In the last 40\%, both models are wrong. As shown in Fig.~\ref{fig:downsampled}, the image is extremely blurry, so the user must rely on the models to inform their prediction. When offered model predictions, in this survey, \textbf{255} of \textbf{600} responses are correct (42.5\% accuracy), a 15.3 point improvement over no model guidance. We observe that humans trust NBDT-explained prediction more often than the saliency-explained predictions. Out of \textbf{600} responses, \textbf{312} responses agreed with the NBDT’s prediction, \textbf{167} responses agreed with the base model’s prediction, and \textbf{119} responses disagreed with both model’s predictions. Note that a majority of user decisions ($\sim 80\%$) agreed with either model prediction, even though neither model prediction was correct in 40\% of examples, showing our images were sufficiently blurred to force reliance on the models. Furthermore, 52\% of responses agreed with NBDT (against saliency's 28\%), even though only 30\% of NBDT predictions were correct, showing improvement in model trust.

\subsection{Survey: Human-Diagnosed Level of Trust}
\label{sec:human-trust}

The explanation of an NBDT prediction is the visualization of the path traversed. We then compare these NBDT explanations to other explainability methods in human studies.
Specifically, we ask participants to pick an expert to trust (Appendix, Figure \ref{fig:survey}), based on the expert's explanation -- a saliency map (ResNet18, GradCAM), a decision tree (NBDT), or neither. We only use samples where ResNet18 and NBDT predictions agree. Of 374 respondents that picked one method over the other, \textbf{65.9\%} prefer NBDT explanations; for misclassified samples, \textbf{73.5\%} prefer NBDT. This supports the previous survey's results, showing humans trust NBDTs more than current saliency techniques when explicitly asked.

\subsection{Analysis: Identifying Faulty Dataset Labels}
\label{sec:dataset-debugging}
\begin{figure}[t]
    \centering
    \includegraphics[width=\textwidth]{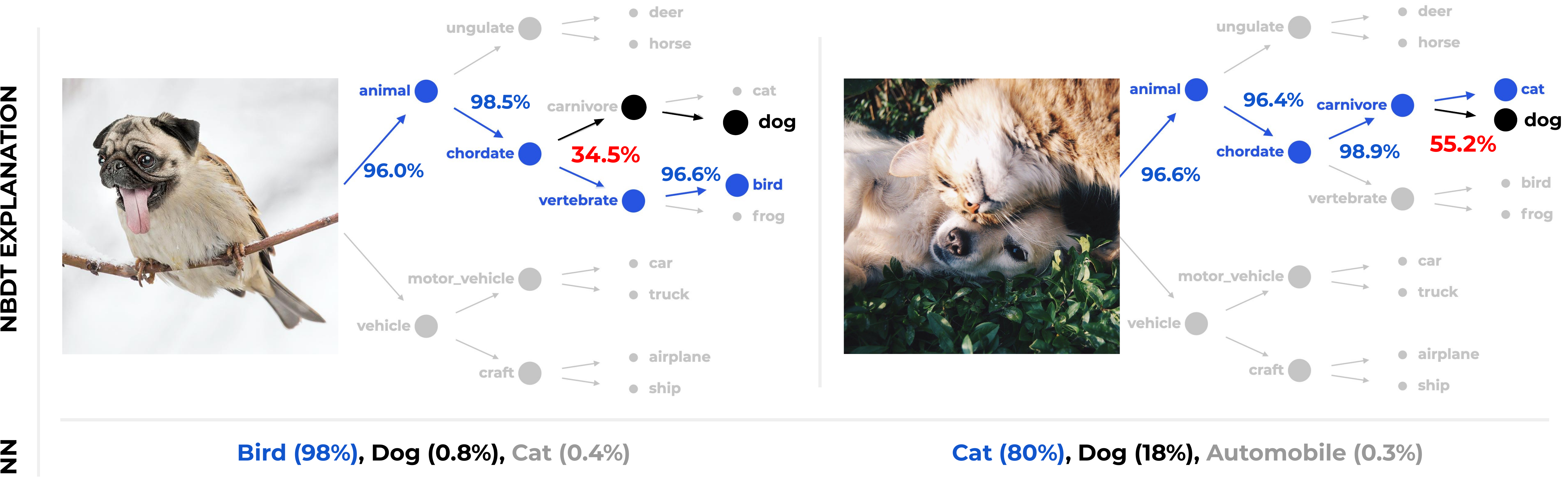}
    \caption{\textbf{Types of Ambiguous Labels.} All these examples have ambiguous labels. With NBDT (top), the decision rule deciding between equally-plausible classes has low certainty (red, 30-50\%). All other decision rules have high certainty (blue, 96\%+). The juxtaposition of high and low certainty decision rules makes ambiguous labels easy to distinguish. By contrast, ResNet18 (bottom) still picks one class with high probability. (Left) An extreme example of a ``spug'' that may plausibly belong to two classes. (Right) Image containing two animals of different classes. Photo ownership: ``Spug'' by Arne Fredriksen at gyyporama.com. Used with permission. Second image is CC-0 licensed at pexels.com.}
    \label{fig:ambiguous_labels_types}
\end{figure}

There are several types of ambiguous labels (Figure \ref{fig:ambiguous_labels_types}), any of which could hurt model performance for an image classification dataset like ImageNet. To find these images, we use entropy in NBDT decisions, which we find is a much stronger indicator of ambiguity than entropy in the original neural network prediction. The intuition is as follows: If all intermediate decisions have high certainty except for a few decisions, those decisions are deciding between multiple equally plausible cases. Using this intuition, we can identify ambiguous labels by finding samples with high ``path entropy'' -- or highly disparate entropies for intermediate decisions on the NBDT prediction path.

\begin{figure}[t]
    \centering
    \includegraphics[width=\textwidth]{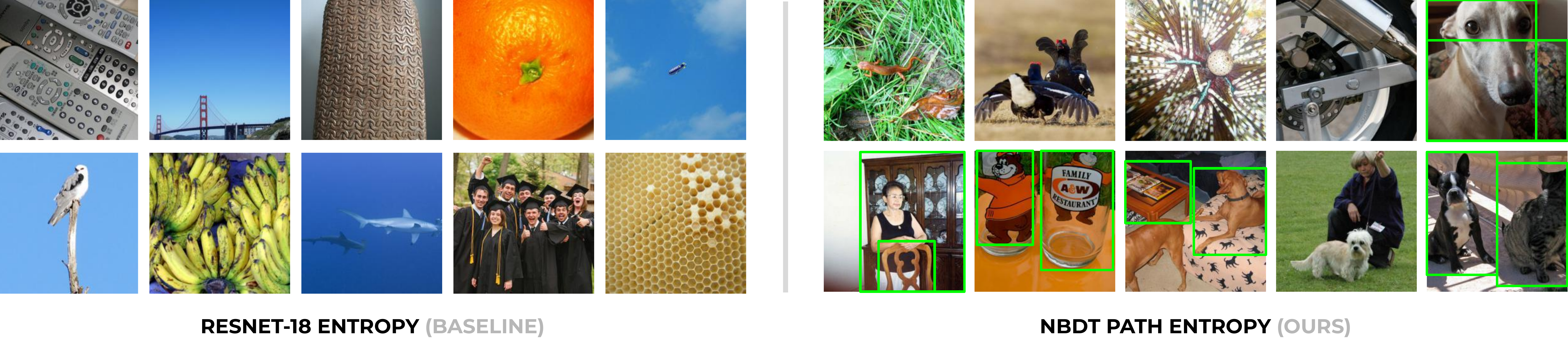}
    \caption{\textbf{ImageNet Ambiguous Labels.} These images suggest that NBDT path entropy uniquely identifies ambiguous labels in Imagenet, without object detection labels. We plot ImageNet validation samples that induce the most 2-class confusion, using TinyImagenet200-trained models. Note that ImageNet classes do not include people. (Left) Run ResNet18 and find samples that (a) maximize entropy between the top 2 classes and (b) minimize entropy across all classes, where the top 2 classes are averaged. Despite high model uncertainty, half the classes are from the training set -- bee, orange, bridge, banana, remote control -- and do not show visual ambiguity. (Right) For NBDT, compute entropy for each node's predicted distribution; take the difference between the largest and smallest values. Now, half of the images contain truly ambiguous content for a classifier; we draw green boxes around pairs of objects that could each plausibly be used for the image class.}
    \label{fig:ambiguous_labels_imagenet}
\end{figure}

Per Figure \ref{fig:ambiguous_labels_imagenet}, the highest ``path entropy'' samples in ImageNet contain multiple objects, where each object could plausibly be used for the image class. In contrast, samples that induce the highest entropy in the baseline neural network do not suggest ambiguous labels. This suggests NBDT entropy is more informative compared to that of a standard neural network.

\section{Conclusion}
In this work, we propose Neural-Backed Decision Trees that see (1) improved accuracy: NBDTs out-generalize (16\%+), improve (2\%+), and match (0.15\%) or outperform (1\%+) state-of-the-art neural networks on CIFAR10, CIFAR100, TinyImageNet, and ImageNet. We also show (2) improved interpretability by drawing unique insights from our hierarchy, confirming that humans trust NBDT's over saliency and illustrate how path entropy can be used to identify ambiguous labels.  This challenges the conventional supposition of a dichotomy between accuracy and interpretability, paving the way for jointly accurate \textit{and} interpretable models in real-world deployments.

\newpage
\clearpage

\bibliography{iclr2021_conference}

\begin{thebibliography}{56}
\providecommand{\natexlab}[1]{#1}
\providecommand{\url}[1]{\texttt{#1}}
\expandafter\ifx\csname urlstyle\endcsname\relax
  \providecommand{\doi}[1]{doi: #1}\else
  \providecommand{\doi}{doi: \begingroup \urlstyle{rm}\Url}\fi

\bibitem[Ahmed et~al.(2016)Ahmed, Baig, and Torresani]{network_of_experts}
Karim Ahmed, Mohammadharis Baig, and Lorenzo Torresani.
\newblock Network of experts for large-scale image categorization.
\newblock volume 9911, April 2016.

\bibitem[Alaniz \& Akata(2019)Alaniz and Akata]{xoc}
Stephan Alaniz and Zeynep Akata.
\newblock {XOC:} explainable observer-classifier for explainable binary
  decisions.
\newblock \emph{CoRR}, abs/1902.01780, 2019.

\bibitem[Baek et~al.(2017)Baek, Kim, and Kim]{decision_jungle}
Seungryul Baek, Kwang~In Kim, and Tae{-}Kyun Kim.
\newblock Deep convolutional decision jungle for image classification.
\newblock \emph{CoRR}, abs/1706.02003, 2017.

\bibitem[Banerjee(1990)]{1990_dt_init}
Arunava Banerjee.
\newblock Initializing neural networks using decision trees.
\newblock 1990.

\bibitem[Banerjee(1994)]{1994_dt_init}
Arunava Banerjee.
\newblock Initializing neural networks using decision trees.
\newblock In \emph{Proceedings of the International Workshop on Computational
  Learning and Natural Learning Systems}, pp.\  3--15. MIT Press, 1994.

\bibitem[Bi{\c{c}}ici et~al.(2018)Bi{\c{c}}ici, Keskin, and Akarun]{cign}
Ufuk~Can Bi{\c{c}}ici, Cem Keskin, and Lale Akarun.
\newblock Conditional information gain networks.
\newblock In \emph{2018 24th International Conference on Pattern Recognition
  (ICPR)}, pp.\  1390--1395. IEEE, 2018.

\bibitem[Boz(2000)]{dectext}
Olcay Boz.
\newblock Converting a trained neural network to a decision tree dectext -
  decision tree extractor.
\newblock In \emph{ICMLA}, 2000.

\bibitem[Brust \& Denzler(2019)Brust and Denzler]{brust2019integrating}
Clemens-Alexander Brust and Joachim Denzler.
\newblock Integrating domain knowledge: using hierarchies to improve deep
  classifiers.
\newblock In \emph{Asian Conference on Pattern Recognition}, pp.\  3--16.
  Springer, 2019.

\bibitem[Carvalho et~al.(2019)Carvalho, Pereira, and
  Cardoso]{carvalho2019machine}
Diogo~V Carvalho, Eduardo~M Pereira, and Jaime~S Cardoso.
\newblock Machine learning interpretability: A survey on methods and metrics.
\newblock \emph{Electronics}, 8\penalty0 (8):\penalty0 832, 2019.

\bibitem[Craven \& Shavlik(1996)Craven and Shavlik]{craven1996extracting}
Mark Craven and Jude~W Shavlik.
\newblock Extracting tree-structured representations of trained networks.
\newblock In \emph{Advances in neural information processing systems}, pp.\
  24--30, 1996.

\bibitem[Craven \& Shavlik(1994)Craven and Shavlik]{craven1994using}
Mark~W Craven and Jude~W Shavlik.
\newblock Using sampling and queries to extract rules from trained neural
  networks.
\newblock In \emph{Machine learning proceedings 1994}, pp.\  37--45. Elsevier,
  1994.

\bibitem[Dancey et~al.(2004)Dancey, McLean, and
  Bandar]{decision_tree_extraction}
Darren Dancey, David McLean, and Zuhair Bandar.
\newblock Decision tree extraction from trained neural networks.
\newblock January 2004.

\bibitem[Deng et~al.(2009)Deng, Dong, Socher, Li, Li, and Fei-Fei]{imagenet}
J.~Deng, W.~Dong, R.~Socher, L.-J. Li, K.~Li, and L.~Fei-Fei.
\newblock {ImageNet: A Large-Scale Hierarchical Image Database}.
\newblock In \emph{CVPR09}, 2009.

\bibitem[Deng et~al.()Deng, Ding, Jia, Frome, Murphy, Bengio, Li, Neven, and
  Adam]{denglarge}
Jia Deng, Nan Ding, Yangqing Jia, Andrea Frome, Kevin Murphy, Samy Bengio, Yuan
  Li, Hartmut Neven, and Hartwig Adam.
\newblock Large-scale object classification using label relation graphs.

\bibitem[Deng et~al.(2012)Deng, Krause, Berg, and Fei-Fei]{hedging}
Jia Deng, Jonathan Krause, Alexander~C Berg, and Li~Fei-Fei.
\newblock Hedging your bets: Optimizing accuracy-specificity trade-offs in
  large scale visual recognition.
\newblock In \emph{2012 IEEE Conference on Computer Vision and Pattern
  Recognition}, pp.\  3450--3457. IEEE, 2012.

\bibitem[Doshi-Velez \& Kim(2017)Doshi-Velez and Kim]{doshi2017towards}
Finale Doshi-Velez and Been Kim.
\newblock Towards a rigorous science of interpretable machine learning.
\newblock \emph{arXiv preprint arXiv:1702.08608}, 2017.

\bibitem[Dua \& Graff(2017)Dua and Graff]{uci}
Dheeru Dua and Casey Graff.
\newblock {UCI} machine learning repository, 2017.
\newblock URL \url{http://archive.ics.uci.edu/ml}.

\bibitem[Frosst \& Hinton(2017)Frosst and Hinton]{soft_decision_tree}
Nicholas Frosst and Geoffrey~E. Hinton.
\newblock Distilling a neural network into a soft decision tree.
\newblock \emph{CoRR}, abs/1711.09784, 2017.

\bibitem[Guo et~al.(2018)Guo, Liu, Bakker, Guo, and Lew]{cnnrnn}
Yanming Guo, Yu~Liu, Erwin~M Bakker, Yuanhao Guo, and Michael~S Lew.
\newblock Cnn-rnn: a large-scale hierarchical image classification framework.
\newblock \emph{Multimedia Tools and Applications}, 77\penalty0 (8):\penalty0
  10251--10271, 2018.

\bibitem[Hinton et~al.(2015)Hinton, Vinyals, and Dean]{hinton2015distilling}
Geoffrey Hinton, Oriol Vinyals, and Jeff Dean.
\newblock Distilling the knowledge in a neural network.
\newblock \emph{arXiv preprint arXiv:1503.02531}, 2015.

\bibitem[Humbird et~al.(2018)Humbird, Peterson, and McClarren]{2018_dt_init}
Kelli Humbird, Luc Peterson, and Ryan McClarren.
\newblock Deep neural network initialization with decision trees.
\newblock \emph{IEEE Transactions on Neural Networks and Learning Systems},
  PP:\penalty0 1--10, October 2018.

\bibitem[Ivanova \& Kubat(1995{\natexlab{a}})Ivanova and Kubat]{1995_dt_init}
Irena Ivanova and Miroslav Kubat.
\newblock Initialization of neural networks by means of decision trees.
\newblock \emph{Knowledge-Based Systems}, 8\penalty0 (6):\penalty0 333 -- 344,
  1995{\natexlab{a}}.
\newblock Knowledge-based neural networks.

\bibitem[Ivanova \& Kubat(1995{\natexlab{b}})Ivanova and Kubat]{1995_dt_nn}
Irena Ivanova and Miroslav Kubat.
\newblock Decision-tree based neural network (extended abstract).
\newblock In \emph{Machine Learning: ECML-95}, pp.\  295--298, Berlin,
  Heidelberg, 1995{\natexlab{b}}. Springer Berlin Heidelberg.

\bibitem[Keskin \& Izadi(2018)Keskin and Izadi]{keskin2018splinenets}
Cem Keskin and Shahram Izadi.
\newblock Splinenets: Continuous neural decision graphs.
\newblock In \emph{Advances in Neural Information Processing Systems}, pp.\
  1994--2004, 2018.

\bibitem[Kontschieder et~al.(2015)Kontschieder, Fiterau, Criminisi, and
  Rota~Bulo]{deep_neural_decision_forests}
Peter Kontschieder, Madalina Fiterau, Antonio Criminisi, and Samuel Rota~Bulo.
\newblock Deep neural decision forests.
\newblock In \emph{The IEEE International Conference on Computer Vision
  (ICCV)}, December 2015.

\bibitem[Krishnan et~al.(1999)Krishnan, Sivakumar, and
  Bhattacharya]{1999_extract_dt}
R.~Krishnan, G.~Sivakumar, and P.~Bhattacharya.
\newblock Extracting decision trees from trained neural networks.
\newblock \emph{Pattern Recognition}, 32\penalty0 (12):\penalty0 1999 -- 2009,
  1999.

\bibitem[Krizhevsky(2009)]{cifar}
Alex Krizhevsky.
\newblock Learning multiple layers of features from tiny images.
\newblock Technical report, 2009.

\bibitem[Le \& Yang(2015)Le and Yang]{Le2015TinyIV}
Ya~Le and Xuan Yang.
\newblock Tiny imagenet visual recognition challenge.
\newblock 2015.

\bibitem[LeCun et~al.(2010)LeCun, Cortes, and Burges]{lecun2010mnist}
Yann LeCun, Corinna Cortes, and CJ~Burges.
\newblock Mnist handwritten digit database.
\newblock \emph{ATT Labs [Online]. Available: http://yann. lecun.
  com/exdb/mnist}, 2, 2010.

\bibitem[Lipton(2016)]{lipton2016mythos}
Zachary~Chase Lipton.
\newblock The mythos of model interpretability. corr abs/1606.03490 (2016).
\newblock \emph{arXiv preprint arXiv:1606.03490}, 2016.

\bibitem[Lundberg et~al.(2020)Lundberg, Erion, Chen, DeGrave, Prutkin, Nair,
  Katz, Himmelfarb, Bansal, and Lee]{lundberg2020local}
SM~Lundberg, G~Erion, H~Chen, A~DeGrave, JM~Prutkin, B~Nair, R~Katz,
  J~Himmelfarb, N~Bansal, and S-i Lee.
\newblock From local explanations to global understanding with explainable ai
  for trees, nat. mach. intell., 2, 56--67, 2020.

\bibitem[McGill \& Perona(2017)McGill and Perona]{dynamic_routing}
Mason McGill and Pietro Perona.
\newblock Deciding how to decide: Dynamic routing in artificial neural
  networks.
\newblock In \emph{ICML}, 2017.

\bibitem[Mohammed \& Umaashankar(2018)Mohammed and
  Umaashankar]{mohammed2018effectiveness}
Abdul~Arfat Mohammed and Venkatesh Umaashankar.
\newblock Effectiveness of hierarchical softmax in large scale classification
  tasks.
\newblock In \emph{2018 International Conference on Advances in Computing,
  Communications and Informatics (ICACCI)}, pp.\  1090--1094. IEEE, 2018.

\bibitem[Murdock et~al.(2016)Murdock, Li, Zhou, and Duerig]{blockout}
Calvin Murdock, Zhen Li, Howard Zhou, and Tom Duerig.
\newblock Blockout: Dynamic model selection for hierarchical deep networks.
\newblock In \emph{Proceedings of the IEEE conference on computer vision and
  pattern recognition}, pp.\  2583--2591, 2016.

\bibitem[Murthy et~al.(2016)Murthy, Singh, Chen, Manmatha, and
  Comaniciu]{deep_decision_network}
Venkatesh~N. Murthy, Vivek Singh, Terrence Chen, R.~Manmatha, and Dorin
  Comaniciu.
\newblock Deep decision network for multi-class image classification.
\newblock In \emph{The IEEE Conference on Computer Vision and Pattern
  Recognition (CVPR)}, June 2016.

\bibitem[Petsiuk et~al.(2018)Petsiuk, Das, and Saenko]{rise}
Vitali Petsiuk, Abir Das, and Kate Saenko.
\newblock Rise: Randomized input sampling for explanation of black-box models.
\newblock In \emph{Proceedings of the British Machine Vision Conference
  (BMVC)}, 2018.

\bibitem[Poursabzi-Sangdeh et~al.(2018)Poursabzi-Sangdeh, Goldstein, Hofman,
  Vaughan, and Wallach]{measuringinterp}
F~Poursabzi-Sangdeh, D~Goldstein, J~Hofman, J~Vaughan, and H~Wallach.
\newblock Manipulating and measuring model interpretability.
\newblock In \emph{MLConf}, 2018.

\bibitem[Redmon \& Farhadi(2017)Redmon and Farhadi]{yolo9000}
Joseph Redmon and Ali Farhadi.
\newblock Yolo9000: better, faster, stronger.
\newblock In \emph{Proceedings of the IEEE conference on computer vision and
  pattern recognition}, pp.\  7263--7271, 2017.

\bibitem[Ribeiro et~al.(2016)Ribeiro, Singh, and Guestrin]{lime}
Marco~Tulio Ribeiro, Sameer Singh, and Carlos Guestrin.
\newblock "why should {I} trust you?": Explaining the predictions of any
  classifier.
\newblock In \emph{Proceedings of the 22nd {ACM} {SIGKDD} International
  Conference on Knowledge Discovery and Data Mining, San Francisco, CA, USA,
  August 13-17, 2016}, pp.\  1135--1144, 2016.

\bibitem[Rota~Bulo \& Kontschieder(2014)Rota~Bulo and
  Kontschieder]{neural_decision_forest}
Samuel Rota~Bulo and Peter Kontschieder.
\newblock Neural decision forests for semantic image labelling.
\newblock In \emph{Proceedings of the IEEE Conference on Computer Vision and
  Pattern Recognition}, pp.\  81--88, 2014.

\bibitem[Roy \& Todorovic(2016)Roy and Todorovic]{neural_regression_forest}
Anirban Roy and Sinisa Todorovic.
\newblock Monocular depth estimation using neural regression forest.
\newblock In \emph{Proceedings of the IEEE conference on computer vision and
  pattern recognition}, pp.\  5506--5514, 2016.

\bibitem[Rudin(2018)]{rudin2018stop}
C~Rudin.
\newblock Stop explaining black box machine learning models for high stakes
  decisions and use interpretable models instead. manuscript based on c. rudin
  please stop explaining black box machine learning models for high stakes
  decisions.
\newblock In \emph{Proceedings of NeurIPS 2018 Workshop on Critiquing and
  Correcting Trends in Learning}, 2018.

\bibitem[Selvaraju et~al.(2017)Selvaraju, Cogswell, Das, Vedantam, Parikh, and
  Batra]{gradcam}
Ramprasaath~R Selvaraju, Michael Cogswell, Abhishek Das, Ramakrishna Vedantam,
  Devi Parikh, and Dhruv Batra.
\newblock Grad-cam: Visual explanations from deep networks via gradient-based
  localization.
\newblock In \emph{IEEE Conference on Computer Vision and Pattern Recognition
  (CVPR)}, pp.\  618--626, 2017.

\bibitem[Shazeer et~al.(2017)Shazeer, Mirhoseini, Maziarz, Davis, Le, Hinton,
  and Dean]{shazeer2017outrageously}
Noam Shazeer, Azalia Mirhoseini, Krzysztof Maziarz, Andy Davis, Quoc Le,
  Geoffrey Hinton, and Jeff Dean.
\newblock Outrageously large neural networks: The sparsely-gated
  mixture-of-experts layer.
\newblock \emph{arXiv preprint arXiv:1701.06538}, 2017.

\bibitem[Silla \& Freitas(2011)Silla and Freitas]{silla2011survey}
Carlos~N Silla and Alex~A Freitas.
\newblock A survey of hierarchical classification across different application
  domains.
\newblock \emph{Data Mining and Knowledge Discovery}, 22\penalty0
  (1-2):\penalty0 31--72, 2011.

\bibitem[Simonyan et~al.(2013)Simonyan, Vedaldi, and
  Zisserman]{simonyan2013deep}
Karen Simonyan, Andrea Vedaldi, and Andrew Zisserman.
\newblock Deep inside convolutional networks: Visualising image classification
  models and saliency maps.
\newblock \emph{arXiv preprint arXiv:1312.6034}, 2013.

\bibitem[Siu(2019)]{tree_grad}
Chapman Siu.
\newblock Transferring tree ensembles to neural networks.
\newblock In \emph{Neural Information Processing}, pp.\  471--480, 2019.

\bibitem[Springenberg et~al.(2014)Springenberg, Dosovitskiy, Brox, and
  Riedmiller]{grad}
Jost~Tobias Springenberg, Alexey Dosovitskiy, Thomas Brox, and Martin~A.
  Riedmiller.
\newblock Striving for simplicity: The all convolutional net.
\newblock \emph{CoRR}, abs/1412.6806, 2014.

\bibitem[Sundararajan et~al.(2017)Sundararajan, Taly, and Yan]{ig}
Mukund Sundararajan, Ankur Taly, and Qiqi Yan.
\newblock Axiomatic attribution for deep networks.
\newblock \emph{International Conference on Machine Learning (ICML) 2017},
  2017.

\bibitem[Tanno et~al.(2019)Tanno, Arulkumaran, Alexander, Criminisi, and
  Nori]{adaptive_neural_trees}
Ryutaro Tanno, Kai Arulkumaran, Daniel~C. Alexander, Antonio Criminisi, and
  Aditya Nori.
\newblock Adaptive neural trees, 2019.

\bibitem[Teja~Mullapudi et~al.(2018)Teja~Mullapudi, Mark, Shazeer, and
  Fatahalian]{hydranets}
Ravi Teja~Mullapudi, William~R. Mark, Noam Shazeer, and Kayvon Fatahalian.
\newblock Hydranets: Specialized dynamic architectures for efficient inference.
\newblock In \emph{The IEEE Conference on Computer Vision and Pattern
  Recognition (CVPR)}, June 2018.

\bibitem[Veit \& Belongie(2018)Veit and Belongie]{adaptive_inference_graphs}
Andreas Veit and Serge Belongie.
\newblock Convolutional networks with adaptive inference graphs.
\newblock In \emph{The European Conference on Computer Vision (ECCV)},
  September 2018.

\bibitem[Wu et~al.(2017)Wu, Hughes, Parbhoo, and
  Doshi-Velez]{treeregularization}
Mike Wu, M~Hughes, Sonali Parbhoo, and F~Doshi-Velez.
\newblock Beyond sparsity: Tree-based regularization of deep models for
  interpretability.
\newblock In \emph{In: Neural Information Processing Systems (NIPS) Conference.
  Transparent and Interpretable Machine Learning in Safety Critical
  Environments (TIML) Workshop}, 2017.

\bibitem[Yang et~al.(2019)Yang, Bender, Le, and Ngiam]{yang2019condconv}
Brandon Yang, Gabriel Bender, Quoc~V Le, and Jiquan Ngiam.
\newblock Condconv: Conditionally parameterized convolutions for efficient
  inference.
\newblock In \emph{Advances in Neural Information Processing Systems}, pp.\
  1307--1318, 2019.

\bibitem[Zeiler \& Fergus(2014)Zeiler and Fergus]{deconv}
Matthew~D Zeiler and Rob Fergus.
\newblock Visualizing and understanding convolutional networks.
\newblock In \emph{{European Conference on Computer Vision (ECCV)}}, pp.\
  818--833. Springer, 2014.

\bibitem[Zhang et~al.(2016)Zhang, Lin, Brandt, Shen, and
  Sclaroff]{zhang2016top}
Jianming Zhang, Zhe Lin, Jonathan Brandt, Xiaohui Shen, and Stan Sclaroff.
\newblock Top-down neural attention by excitation backprop.
\newblock In \emph{{European Conference on Computer Vision (ECCV)}}, pp.\
  543--559. Springer, 2016.

\end{thebibliography}
\bibliographystyle{iclr2021_conference}

\newpage
\appendix

\section{Acknowledgments}

In addition to NSF CISE Expeditions Award CCF-1730628, UC Berkeley research is supported by gifts from Alibaba, Amazon Web Services, Ant Financial, CapitalOne, Ericsson, Facebook, Futurewei, Google, Intel, Microsoft, Nvidia, Scotiabank, Splunk and VMware. This material is based upon work supported by the National Science Foundation Graduate Research Fellowship under Grant No. DGE 1752814.

\section{Explainability}

\begin{figure}[t]
    \centering
    \begin{subfigure}{0.48\textwidth}
        \centering
        \includegraphics[width=\linewidth,trim={.2cm 9cm 9.5cm 0.9cm},clip]{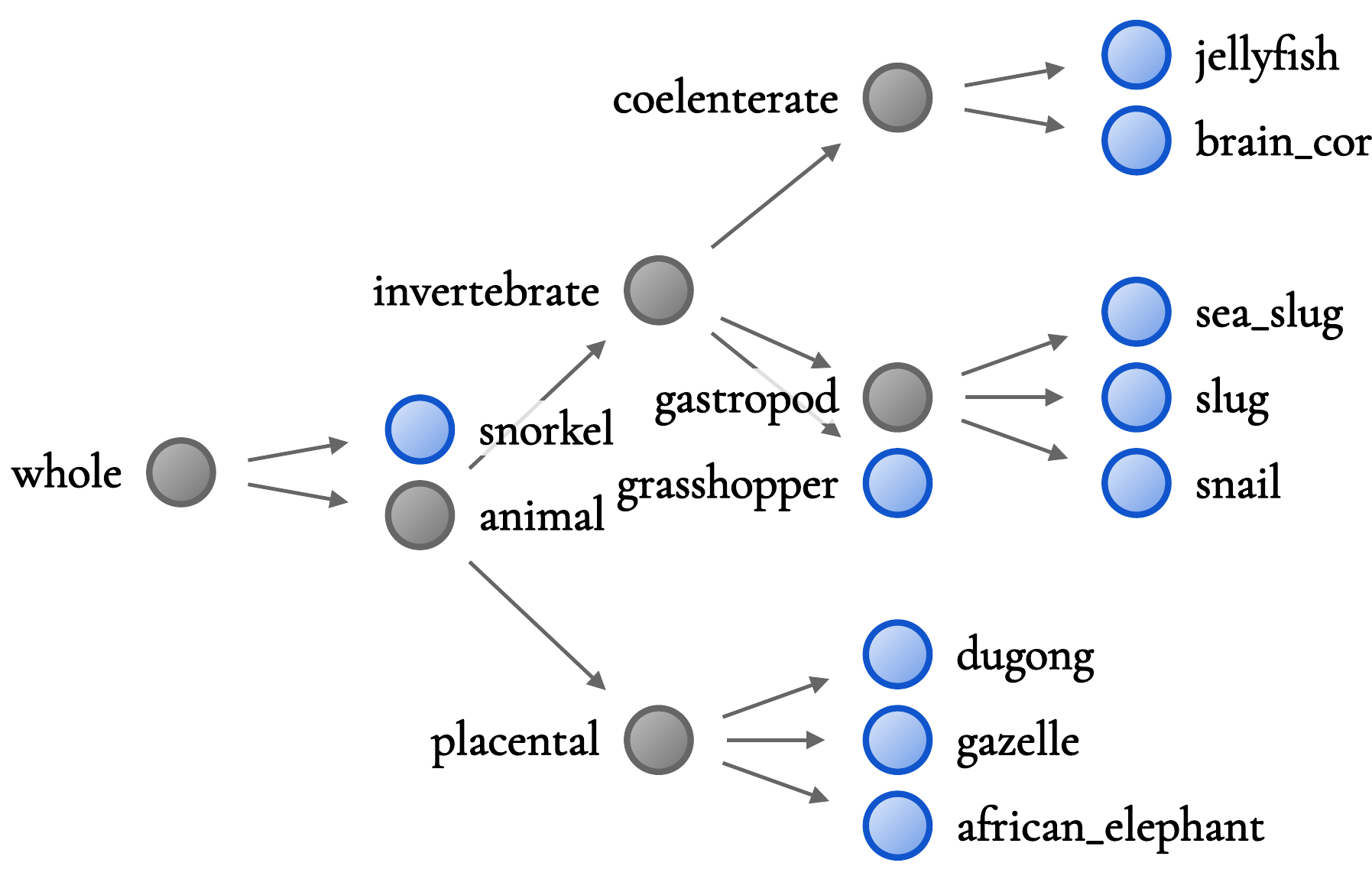}
        \vspace{3pt}
        \caption{\textbf{WordNet Hierarchy}}
        \label{fig:aquatic_wordnet}
    \end{subfigure}
    ~
    \begin{subfigure}{0.48\textwidth}
        \centering
        \includegraphics[width=\linewidth,trim={1cm 9cm 7.5cm 0},clip]{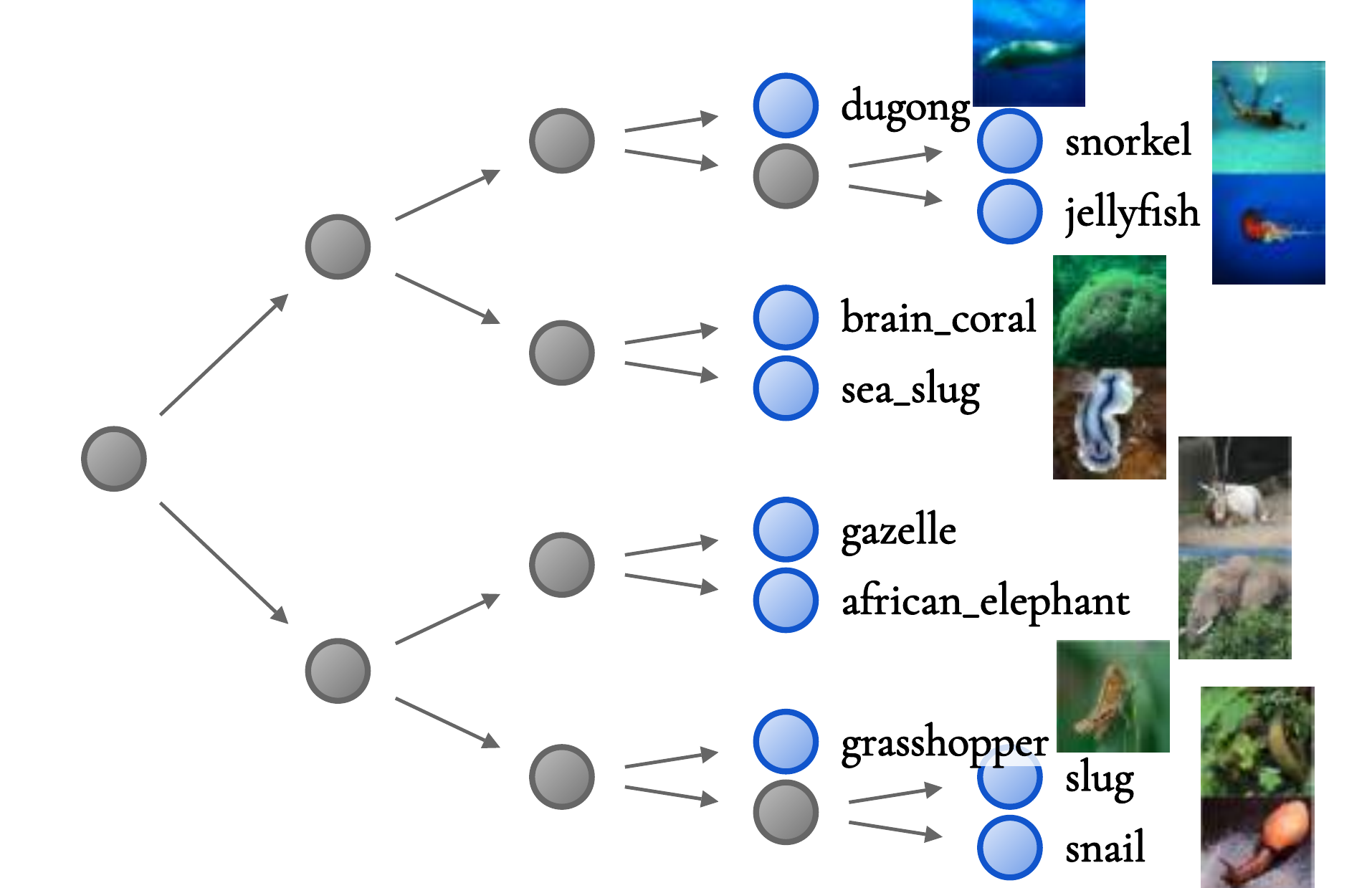}
        \vspace{3pt}
        \caption{\textbf{Induced Hierarchy}}
        \label{fig:aquatic_induced}
    \end{subfigure}
\end{figure}
\label{sec:wordnet}

In this section, we expand on details for interpretability as presented in the original paper, with an emphasis on qualitative use of the hierarchy.

\begin{figure}[t]
    \centering
    \includegraphics[width=\textwidth]{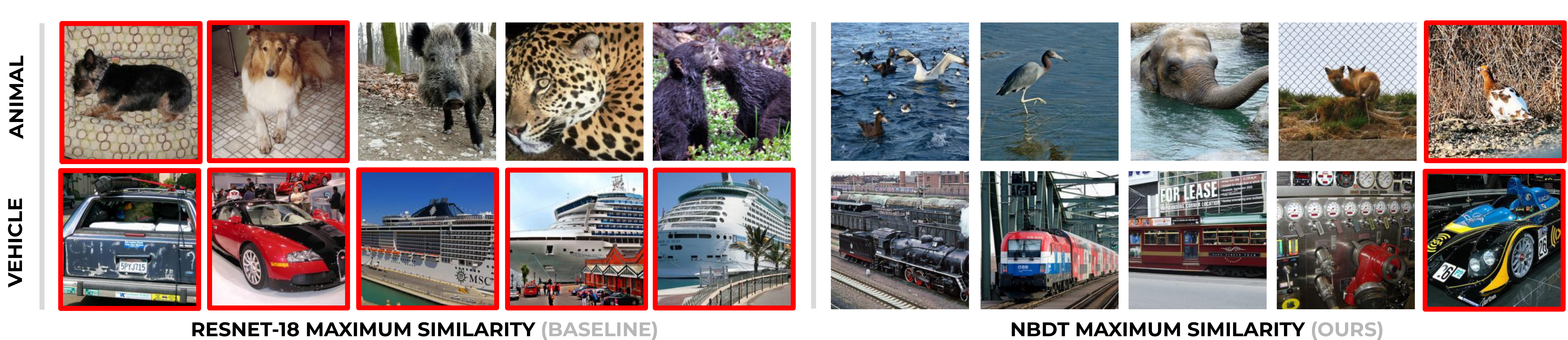}
    \caption{\textbf{Maximum Similarity Examples.} We run two CIFAR10-trained models, one trained with tree supervision loss (NBDT) and one without tree supervision loss (ResNet18). We compute the induced hierarchy of both models and find samples most similar to the \textit{Animal}, and \textit{Motor Vehicle} concepts. Each row represents an inner node, and the red borders indicate images that contain CIFAR10 classes. (1) Note that NBDT's concept of an animal includes classes and contexts it was not trained on; aquatic animals (top-right) and trains (bottom-right) are not a part of CIFAR10. In contrast, ResNet18 largely finds examples closely related to existing CIFAR10 classes (dog, car, boat). This is qualitative evidence that NBDTs better generalize.}
    \label{fig:maximum_similarity_examples}
\end{figure}

\subsection{Maximum Similarity Examples to Visualize Generalization}
\label{sec:visualize-decision-node}

We (1) visually confirm the hypothesized meaning of each node by identifying the most ``representative'' samples, and (2) check that these ``representative'' samples represent that category (e.g., \textit{Animal}) and not just the training classes under that category. We define ``representative'' samples, or maximum similarity examples, to be samples with embeddings most similar to an inner node's representative. We visualize these examples for a model before and after the tree supervision loss (NBDT and ResNet18, respectively). The models are trained on CIFAR10, but samples are drawn from ImageNet. We observe that maximum similarity examples for NBDT contain more unseen classes than ResNet18 (Figure \ref{fig:maximum_similarity_examples}). This suggests that our NBDT is better able to capture high-level concepts such as \textit{Animal}, which is quantitatively confirmed by the superclass evaluation in Table \ref{tab:generalization}.

\subsection{Explainability of Nodes' Visual Meanings}
\label{sec:explainability-node}

\begin{figure}[t]
    \centering
    \begin{subfigure}{0.5\textwidth}
        \centering
        \includegraphics[width=\linewidth,trim={2.5cm 0cm 1.5cm .5cm},clip]{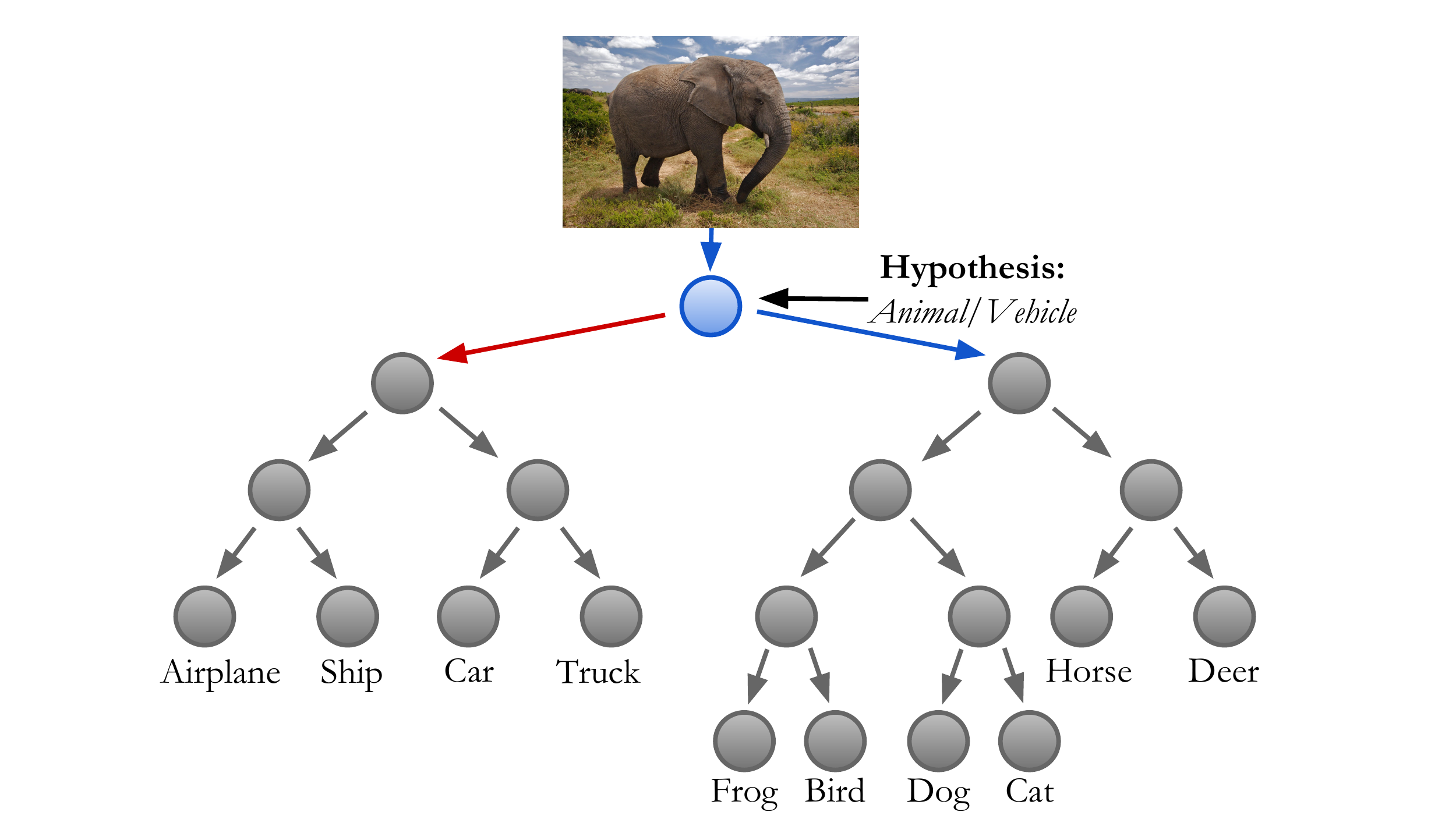}
        \caption{}
        \label{fig:ood_vis}
    \end{subfigure}
    ~
    \begin{subfigure}{0.35\textwidth}
        \centering
        \includegraphics[width=\linewidth]{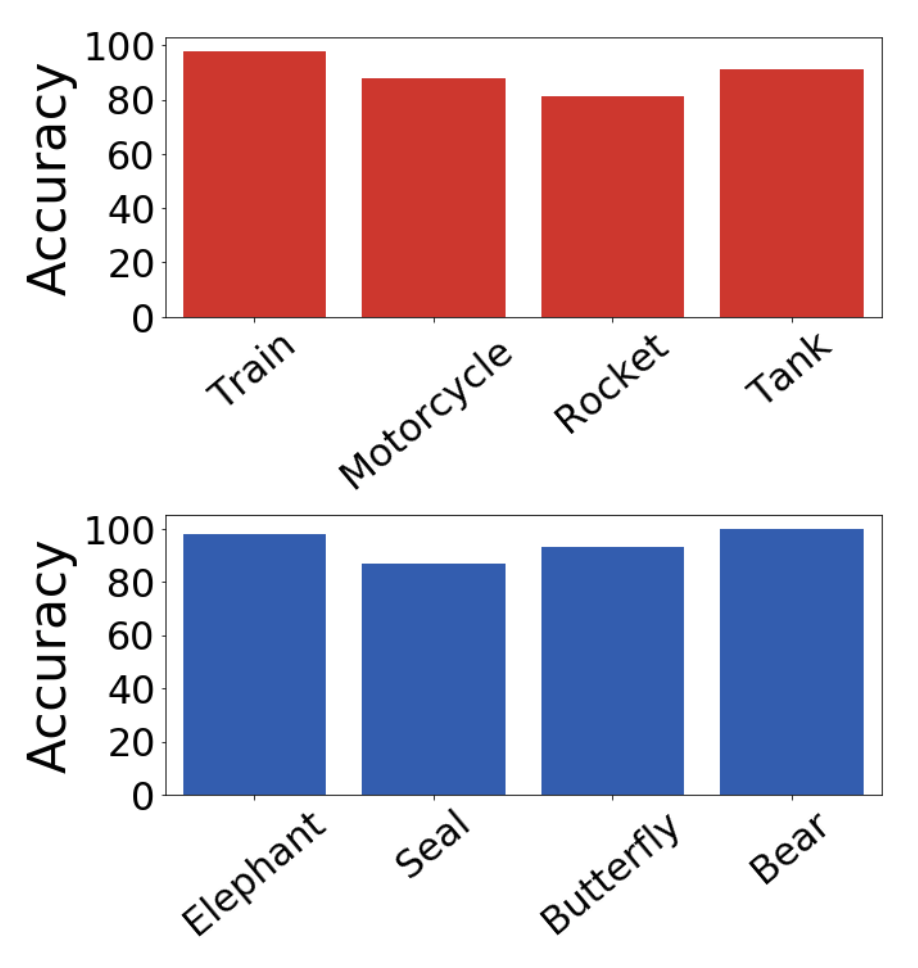}
        \caption{}
        \label{fig:ood_hist}
    \end{subfigure}
    \caption{\textbf{A Node's meaning.} \textbf{(Left)} Visualization of node hypothesis test performed on a CIFAR10-trained WideResNet28x10 model, by sampling from CIFAR100 validation set for OOD classes. \textbf{(Right)} Classification accuracy is high (80-95\%) given unseen CIFAR100 samples of \textit{Vehicles} (top) and \textit{Animals} (bottom), for the WordNet-hypothesized \textit{Animal/Vehicle} node.}
    \label{fig:ood}
\end{figure}

This section describes the method used in Table \ref{tab:generalization} in more detail. Since the induced hierarchy is constructed using model weights, the intermediate nodes are not forced to split on foreground objects. While hierarchies like WordNet provide hypotheses for a node's meaning, the tree may split on unexpected contextual and visual attributes such as \textit{underwater} and \textit{on land}, depicted in Figure 7b. To diagnose a node's visual meaning, we perform the following 4-step test:

\vspace{-1mm}
\begin{enumerate}[leftmargin=6mm]
    \itemsep0.1em
    \item Posit a hypothesis for the node's meaning (e.g. \textit{Animal} \vs \textit{Vehicle}). This hypothesis can be computed automatically from a given taxonomy or deduced from manual inspection of each child's leaves (Figure \ref{fig:ood}).
    \item Collect a dataset with new, unseen classes that test the hypothesised meaning from step 1 (e.g. \textit{Elephant} is an unseen \textit{Animal}). Samples in this dataset are referred to as out-of-distribution (OOD) samples, as they are drawn from a separate labeled dataset.
    \item Pass samples from this dataset through the node. For each sample, check whether the selected child node agrees with the hypothesis.
    \item The accuracy of the hypothesis is the percentage of samples passed to the correct child. If the accuracy is low, repeat with a different hypothesis. 
\end{enumerate}

Figure~\ref{fig:ood_vis} depicts the CIFAR10 tree induced by a WideResNet28x10 model trained on CIFAR10. The WordNet hypothesis is that the root note splits on \textit{Animal} \vs \textit{Vehicle}. We use the CIFAR100 validation set as out-of-distribution images for \textit{Animal} and \textit{Vehicle} classes that are unseen at training time. We then compute the hypothesis' accuracy. Figure~\ref{fig:ood_hist} shows our hypothesis accurately predicts which child each unseen-class's samples traverse.

\subsection{How Model Accuracy Affects Interpretability}

\begin{figure}[t]
    \centering
    \begin{subfigure}{0.42\textwidth}
        \centering
        \includegraphics[width=\linewidth]{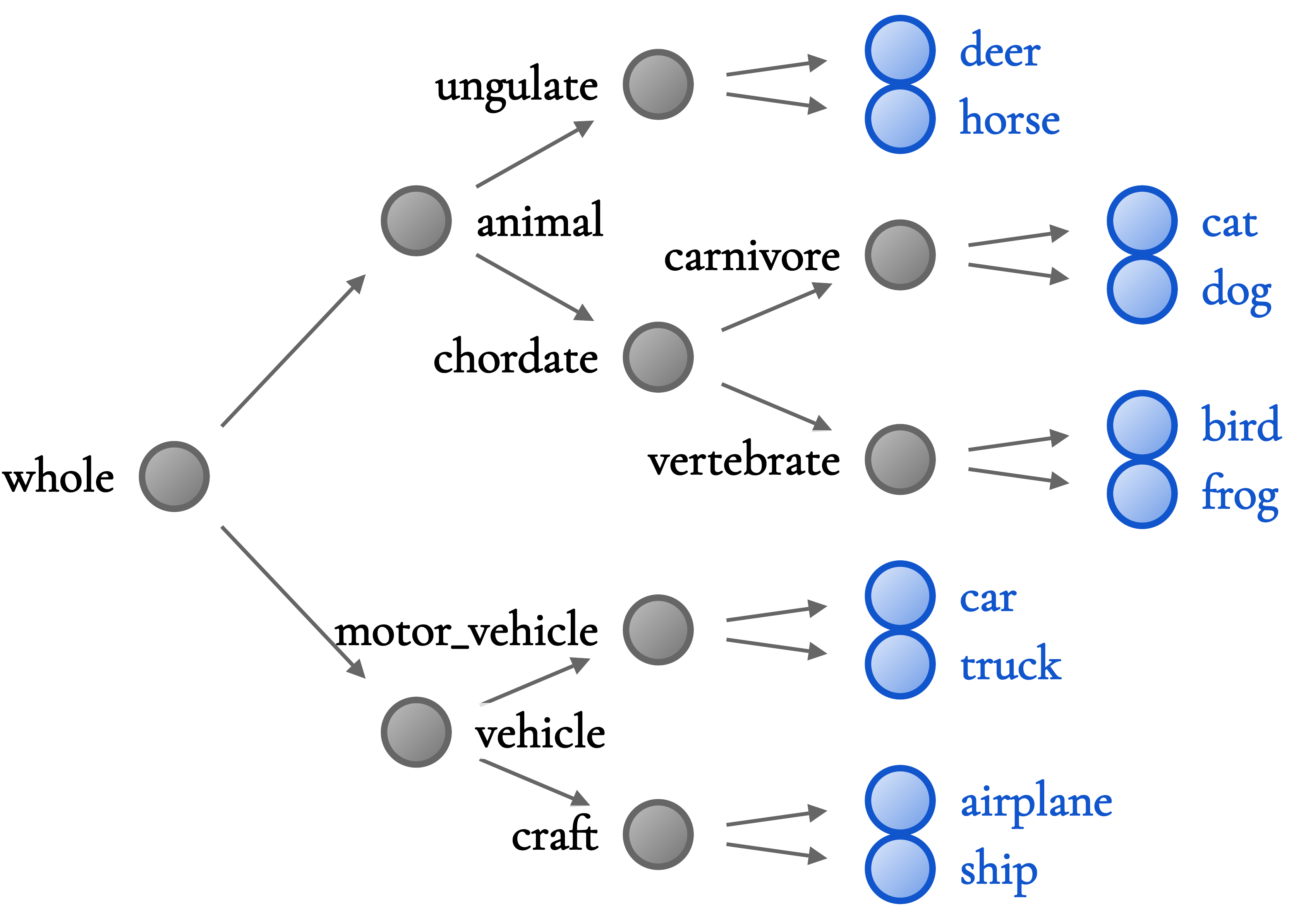}
        \caption{\textbf{WideResNet28x10}}
        \label{fig:induced_tree_wrn}
    \end{subfigure}
    ~
    \begin{subfigure}{0.55\textwidth}
        \centering
        \includegraphics[width=\linewidth,trim={0 3.5cm 0 2.7cm},clip]{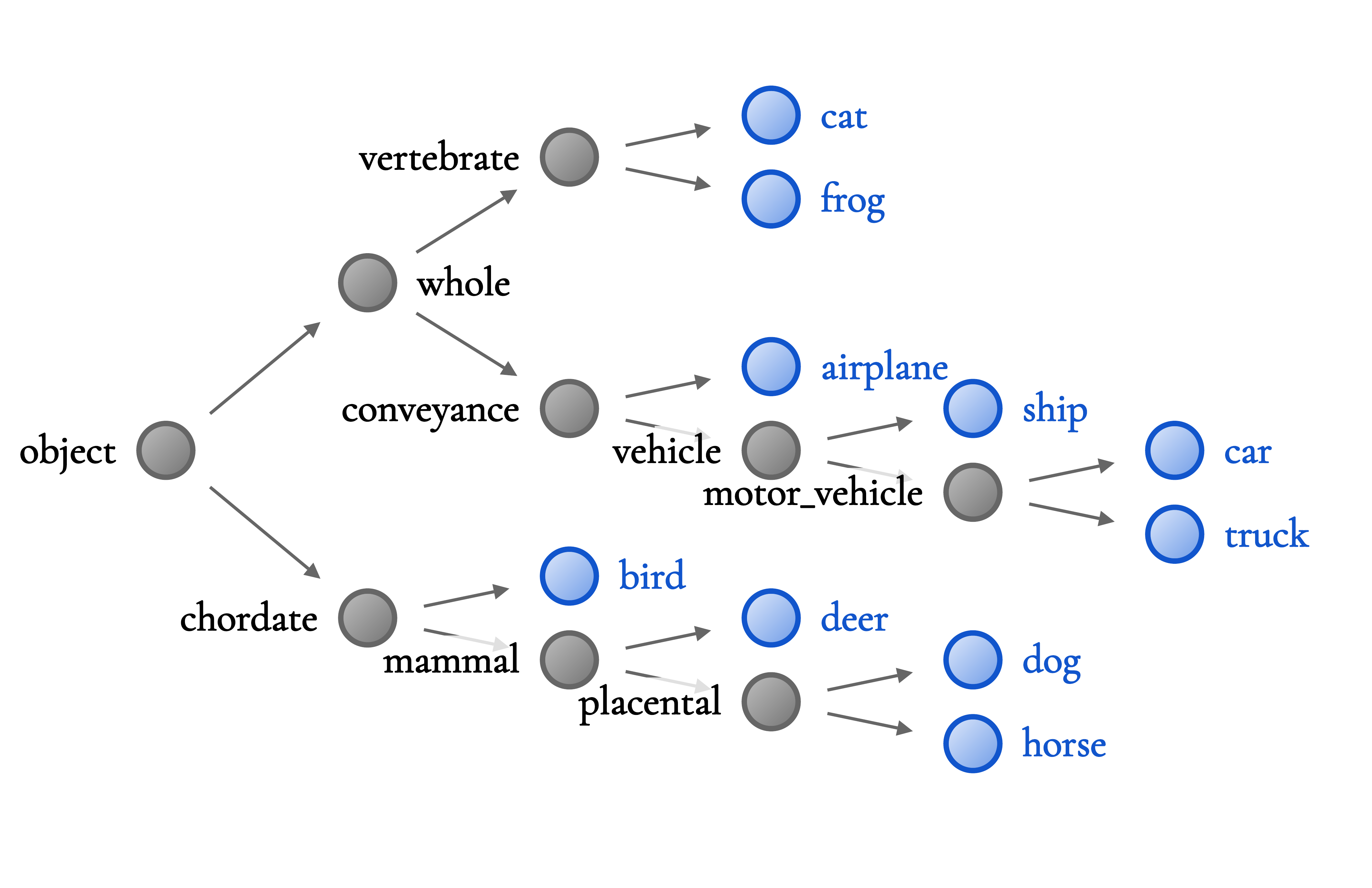}
        \caption{\textbf{ResNet10}}
        \label{fig:induced_tree_rn10}
    \end{subfigure}
    \caption{\textbf{CIFAR10 induced hierarchies}, with automatically-generated WordNet hypotheses for each node. The higher-accuracy (a) WideResNet (97.62\% acc) has a more sensible hierarchy than (b) ResNet's (93.64\% acc): The former groups all \textit{Animal}s together, separate from all \textit{Vehicles}. By contrast, the latter groups \textit{Airplane}, \textit{Cat}, and \textit{Frog}. \color{white}{Easter egg 2!}}
    \label{fig:induced_tree_comp}
\end{figure}

Induced hierarchies are determined by the proximity of class weights, but classes that are close in weight space may not have similar visual meaning: Figure~\ref{fig:induced_tree_comp} depicts the trees induced by WideResNet28x10 and ResNet10, respectively. While the WideResNet induced hierarchy (Figure~\ref{fig:induced_tree_wrn}) groups visually-similar classes, the ResNet (Figure~\ref{fig:induced_tree_rn10}) induced hierarchy does not, grouping classes such as \textit{Frog}, \textit{Cat}, and \textit{Airplane}. This disparity in visual meaning is explained by WideResNet's 4\% higher accuracy: we believe that higher-accuracy models exhibit more visually-sound weight spaces. Thus, unlike previous work, NBDTs feature better interpretability with higher accuracy, instead of sacrificing one for the other. Furthermore, the disparity in hierarchies indicates that a model with low accuracy will not provide interpretable insight into high-accuracy decisions.

\label{sec:model_hierarchies}

\subsection{Visualization of Tree Traversal}

\begin{figure}[t!]
    \begin{subfigure}{0.45\textwidth}
        \centering
        \includegraphics[width=\linewidth]{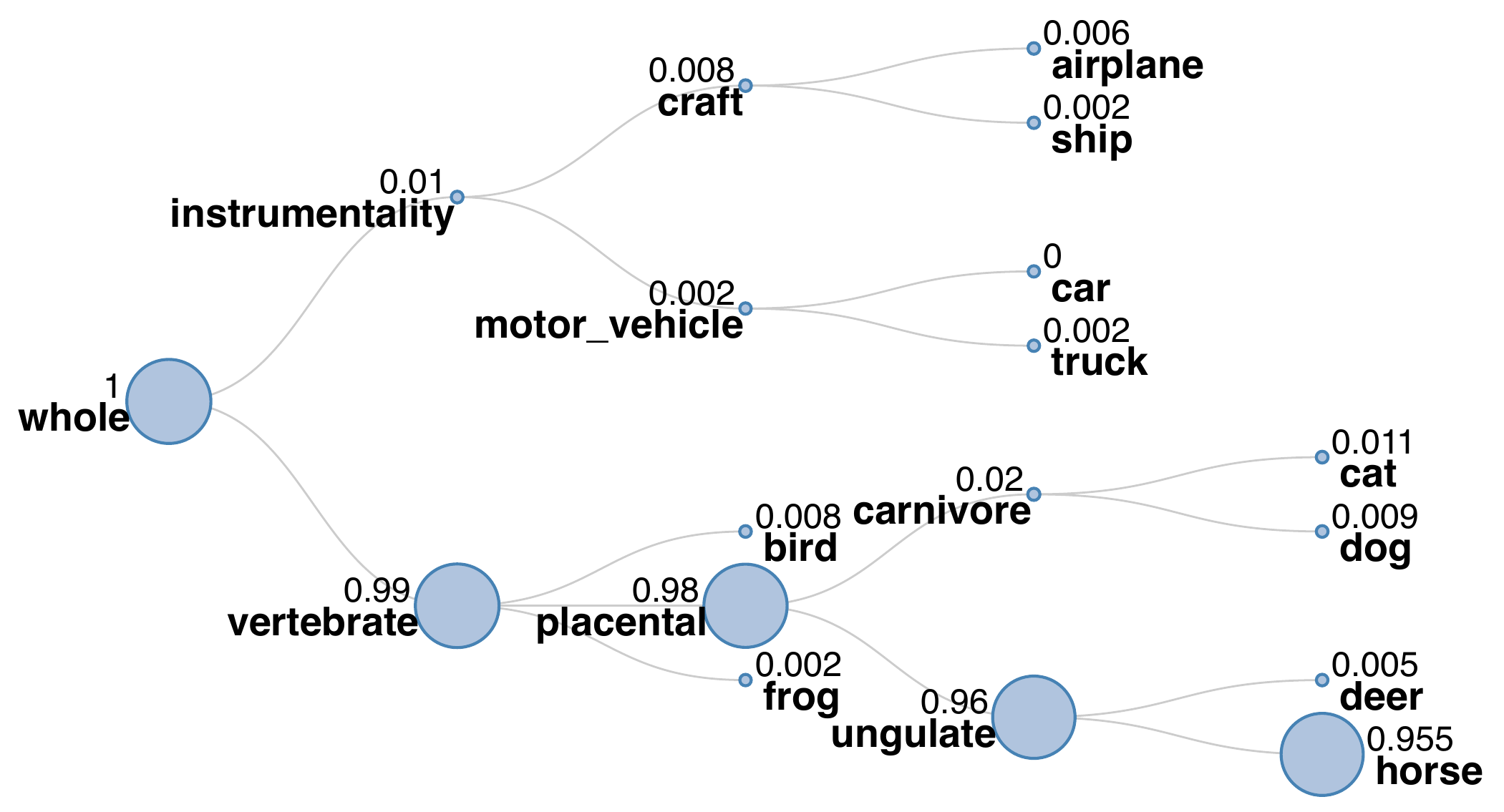}
        \caption{}
        \label{fig:horse_tree}
    \end{subfigure}%
    ~ 
    \begin{subfigure}{0.23\textwidth}
        \centering
        \includegraphics[width=\linewidth]{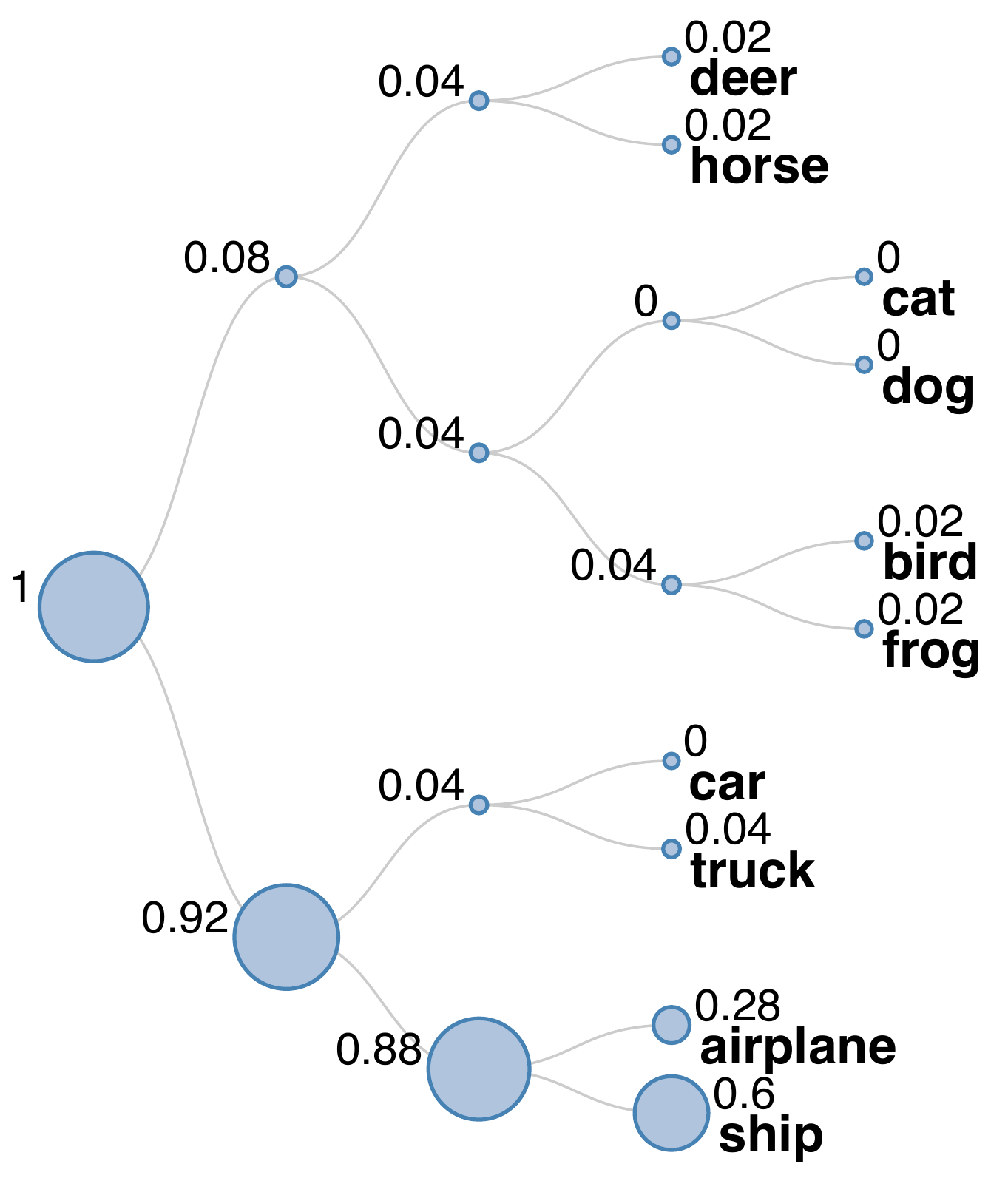}
        \caption{}
        \label{fig:cliff_dwelling_tree}
    \end{subfigure}
        ~ 
    \begin{subfigure}{0.23\textwidth}
        \centering
        \includegraphics[width=\linewidth]{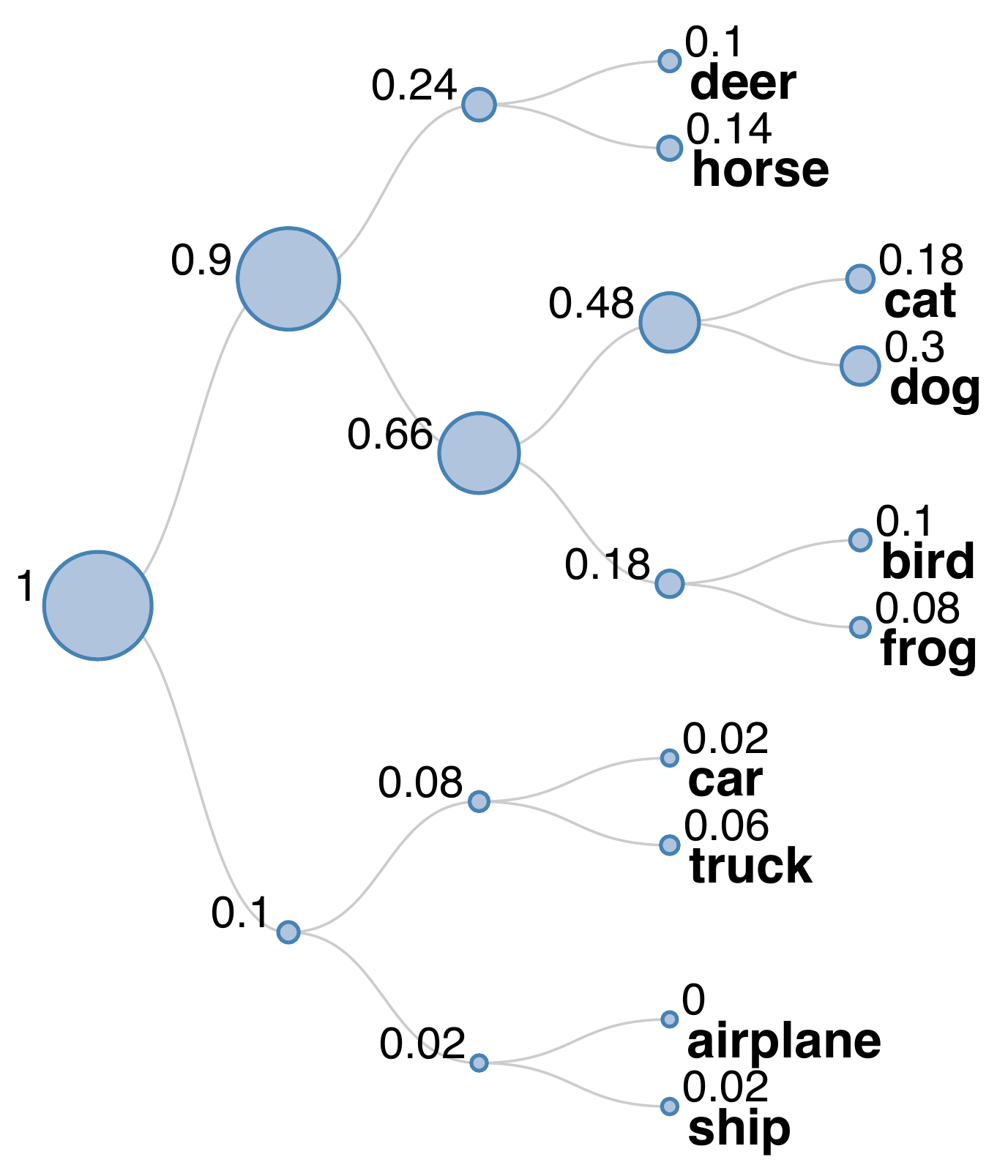}
        \caption{}
        \label{fig:jellyfish_tree}
    \end{subfigure}
    \caption{\textbf{Visualization of path traversal frequency} on an induced hierarchy for CIFAR10. (a) \textbf{In-Distribution: \textit{Horse}} is a training class and thus sees highly focused path traversals. (b) \textbf{Unseen Class: \textit{Seashore}} is largely classified as \textit{Ship} despite not containing any objects, exhibiting model reliance on context (water). (c) \textbf{Unseen Class: \textit{Teddy Bear}} is classified as \textit{Dog}, for sharing visual attributes like color and texture.}
    \label{fig:induced_tree_vis}
\end{figure}

Frequency of path traversals additionally provide insight into general model behavior. Figure~\ref{fig:induced_tree_vis} shows frequency of path traversals for all samples in three classes: a seen class, an unseen class but with seen context, and an unseen class with unseen context.

\textbf{Seen class, seen context}: We visualize tree traversals for all samples in CIFAR10's \textit{Horse} class (Figure~\ref{fig:horse_tree}). As this class is present during training, tree traversal highlights the correct path with extremely high frequency. 
\textbf{Unseen class, seen context}: In Figure~\ref{fig:cliff_dwelling_tree}, we visualize tree traversals for TinyImagenet's \textit{Seashore} class. The model classifies 88\% of \textit{Seashore} samples as ``vehicle with blue context,'' exhibiting reliance on context for decision-making.
\textbf{Unseen class, unseen context}: In Figure~\ref{fig:jellyfish_tree}, we visualize traversals for TinyImagenet's \textit{Teddy Bear}. The model classifies 90\% as \textit{Animal}, belying the model's generalization to stuffed animals. However, the model disperses samples among animals more evenly, with the most furry animal \textit{Dog} receiving the most \textit{Teddy Bear} samples (30\%).

\label{sec:explainability-taxonomy}

\section{Hierarchical Softmax and Conditional Execution}

In the context of neural netework and decision tree hybrids, many works \citep{shazeer2017outrageously,keskin2018splinenets,yang2019condconv,adaptive_neural_trees} leverage conditional execution to improve computational efficiency in a hierarchical classifier. One motivation is to handle large-scale classification problems. 

\subsection{Hard Tree Supervision Loss}
\label{sec:hard-tree-supervision-loss}

\begin{figure}[t]
    \centering
    \includegraphics[width=\textwidth]{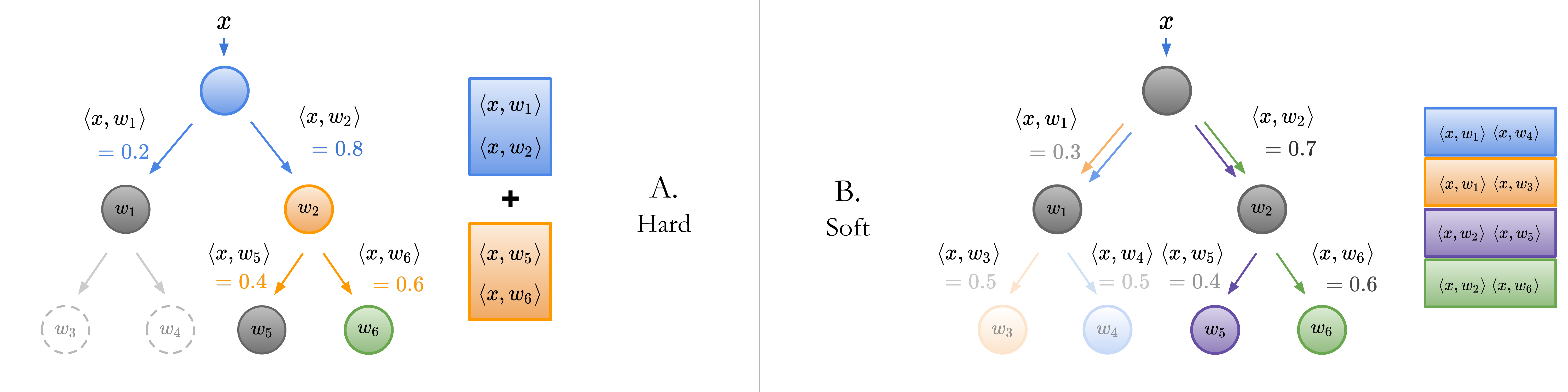}
    \caption{\textbf{Tree Supervision Loss} has two variants: \textbf{Hard Tree Supervision Loss (A)} defines a cross entropy term per node. This is illustrated with the blue box for the blue node and the orange box for the orange node. The cross entropy is taken over the child node probabilities. The green node is the leaf representing a class label. The dotted nodes are not included in the path from the label to the root, so do not have a defined loss. \textbf{Soft Tree Supervision Loss (B)} defines a cross entropy loss over all leaf probabilities. The probability of the green leaf is the product of the probabilities leading up to the root (in this case, $\langle x, w_2 \rangle \langle x, w_6 \rangle = 0.6 \times 0.7$). The probabilities for the other leaves are similarly defined. Each leaf probability is represented with a colored box. The cross entropy is then computed over this leaf probability distribution, represented by the colored box stacked on one another.}
    \label{fig:tsl}
\end{figure}

An alternative loss would be hierarchical softmax -- in other words, one cross entropy loss per decision rule. We denote this the \textit{hard tree supervision loss}, as we construct a variant of hierarchical softmax that (a) supports arbitrary depth trees and (b) is defined over a single, un-augmented fully-connected layer (e.g. $k$-dimensional output for a $k$-leaf tree). The original neural network's loss $\mathcal{L}_\text{original}$ minimizes cross entropy across the classes. For a $k$-class dataset, this is a $k$-way cross entropy loss. Each internal node's goal is similar: minimize cross-entropy loss across the child nodes. For node $i$ with $c$ children, this is a $c$-way cross entropy loss between predicted probabilities $\mathcal{D}(i)_\text{pred}$ and labels $\mathcal{D}(i)_\text{label}$. We refer to this collection of new loss terms as the \textit{hard tree supervision loss} (Eq. \ref{eqn:htsl}). The individual cross entropy losses for each node are scaled so that the original cross entropy loss and the tree supervision loss are weighted equally, by default. 
If we assume $N$ nodes in the tree, excluding leaves, then we would have $N + 1$ different cross entropy loss terms -- the original cross entropy loss and $N$ hard tree supervision loss terms. This is $\mathcal{L}_\text{original} + \mathcal{L}_\text{hard}$, where:

\begin{equation}\label{eqn:htsl}
    \mathcal{L}_\text{hard} = \frac{1}{N} \sum_{i=1}^N \underbrace{\textsc{CrossEntropy}(\mathcal{D}(i)_\text{pred}, \mathcal{D}(i)_\text{label})}_\text{over the $c$ children for each node}.
\end{equation}

\subsection{Hard Inference}
\label{sec:hard-inference}

Hard inference is more intuitive: Starting at the root node, each sample is sent to the child with the most similar representative. We continue picking and traversing the tree until we reach a leaf. The class associated with this leaf is our prediction (Figure~\ref{fig:inference-modes}, A. Hard). More precisely, consider a tree with nodes indexed by $i$ with set of child nodes $C(i)$. Each node $i$ produces a probability of child node $j \in C(i)$; this probability is denoted $p(j|i)$. Each node thus picks the next node using $\text{argmax}_{j \in C(i)} p(j|i)$.

Whereas this inference mode is more intuitive, it underperforms soft inference (Figure \ref{tab:hard-inference}). Furthermore, note that hard tree supervision loss (\ie modified hierarchical softmax) appears to more specifically optimize hard inference. Despite that, hard inference performs worse (Figure \ref{tab:hard-inference-with-hard-loss}) with hard tree supervision loss than the ``soft'' tree supervision loss (Sec \ref{sec:method-supervision}) used in the main paper.

\begin{table}[t]
\small
\centering
\caption{\textbf{Comparisons of Inference Modes} Hard inference performs worse than soft inference. See Table \ref{tab:cifar} in the main manuscript for a comparison against baselines.}

\vspace{5pt}
\begin{tabular*}{\textwidth}{l @{\extracolsep{\fill}} llllll}
\toprule
Method & Backbone     & CIFAR10 & CIFAR100 & TinyImageNet \\
\midrule
NN & WideResNet28x10 & \textit{97.62\%} & \textit{82.09\%}  & \textit{67.65\%} \\
NBDT-H (Ours) & WideResNet28x10 & 97.55\% & 82.21\%  & 64.39\% \\
NBDT-S (Ours) &  WideResNet28x10 & \textbf{97.55\%} & \textbf{82.97\%}  & \textbf{67.72\%}\\

\midrule
NN & ResNet18 & \textit{94.97\%} & \textit{75.92\%}  & \textit{64.13\%}\\
NBDT-H (Ours) & ResNet18     & 94.50\% & 74.29\% & 61.60\%  \\
NBDT-S (Ours) & ResNet18     & \textbf{94.82}\% & \textbf{77.09\%} & \textbf{63.77\%}  \\

\bottomrule
\end{tabular*}
\label{tab:hard-inference}
\end{table}

\begin{table}[t]
\small
\centering
\caption{\textbf{Tree Supervision Loss} Training the NBDT with the tree supervision loss (``TSL'') is superior to (a) training with a hierarchical softmax (``HS'') and to (b) omitting extra loss terms. (``None''). $\Delta$ is the accuracy difference between our soft loss and hierarchical softmax.}
\vspace{5pt}
\begin{tabular*}{\textwidth}{l @{\extracolsep{\fill}} lllllll}
\toprule
Dataset & Backbone     &  NN & Inference & None & TSL & HS & $\Delta$\\
\midrule
CIFAR10 & ResNet18 & 94.97\% & Hard & 94.32\% & 94.50\% & 93.94\% & +0.56\%\\
CIFAR10 & ResNet18 & 94.97\% & Soft & 94.38\% & 94.82\% & 93.97\% & +0.85\% \\
CIFAR100 & ResNet18 & 75.92\% & Hard & 57.63\% & 74.29\% & 73.23\% & +0.94\% \\
CIFAR100 & ResNet18 & 75.92\% & Soft & 61.93\% & 77.09\% & 74.09\% & +1.83\% \\
TinyImageNet & ResNet18 & 64.13\% & Hard & 39.57\% & 61.60\% & 58.89\% & +2.71\%\\
TinyImageNet & ResNet18 & 64.13\% & Soft & 45.51\% & 63.77\% & 61.12\% & +2.65\%\\
\bottomrule
\end{tabular*}
\label{tab:hard-inference-with-hard-loss}
\end{table}

\section{Implementation}

Our inference strategy, as outlined above and in Sec. 3.1 of the paper, includes two phases: (1) featurizing the sample using the neural network backbone and (2) running the embedded decision rules. However, in practice, our inference implementation does not need to run inference with the backbone, separately. In fact, our inference implementation only requires the logits $\hat{y}$ outputted by the network. This is motivated by the knowledge that the average of inner products is equivalent to the inner product of averages. Knowing this, we have the following equivalence, given the fully-connected layer weight matrix $W$, its row vectors $w_i$, featurized sample $x$, and the classes $C$ we are currently interested in.

\begin{equation}
    \langle x, \frac{1}{n}\sum_{i=1}^{|C|} w_i \rangle = \frac{1}{n}\sum_{i=1}^{|C|} \langle x, w_i \rangle = \frac{1}{n}\sum_{i=1}^{|C|} \hat{y}_i, i \in C
\end{equation}

Thus, our inference implementation is simply performed using the logits $\hat{y}$ output by the network.

\section{Experimental Setup}
\label{sec:setup}

To reiterate, our best-performing models for both hard and soft inference were obtained by training with the soft tree supervision loss. All CIFAR10 and CIFAR100 experiments weight the soft loss terms by 1. All TinyImagenet and Imagenet experiments weight the soft loss terms by 10. We found that hard loss performed best when the hard loss weight was $10\times$ that of the corresponding soft loss weight (e.g. weight 10 for CIFAR10, CIFAR100; and weight 100 for TinyImagenet, Imagenet); these hyper-parameters are use for the tree supervision loss comparisons in Table \ref{tab:losses}.

Where possible, we retrain the network from scratch with tree supervision loss. For our remaining training hyperparameters, we largely use default settings found in \url{github.com/kuangliu/pytorch-cifar}: SGD with $0.9$ momentum, $5^{-4}$ weight decay, a starting learning rate of 0.1, decaying by 90\% $\frac{3}{7}$ and $\frac{5}{7}$ of the way through training. We make a few modifications: Training lasts for 200 epochs instead of 350, and we use batch sizes of 512 and 128 on one Titan Xp for CIFAR and TinyImagenet respectively.

In cases where we were unable to reproduce the baseline accuracy (WideResNet), we fine-tuned a pretrained checkpoint with the same settings as above, except with starting learning rate of 0.01.

On Imagenet, we retrain the network from scratch with tree supervision loss. For our remaining hyperparameters, we use settings reported to reproduce EfficientNet-EdgeTPU-Small results at \\ \url{github.com/rwightman/pytorch-image-models}: batch size 128, RMSProp with starting learning rate of 0.064, decaying learning rate by 97\% every 2.4 epochs, weight decay of $10^{-5}$, drop-connect with probability 0.2 on 8 V100s. Our results were obtained with only one model, as opposed to averaging over 8 models, so our reported baseline is 77.23\%, as reported by the EfficientNet authors: \url{https://github.com/tensorflow/tpu/tree/master/models/official/efficientnet/edgetpu#post-training-quantization}.

\begin{figure}[t]
    \centering
    \includegraphics[width=1\textwidth, trim={0cm 10cm 0cm 0cm}, clip]{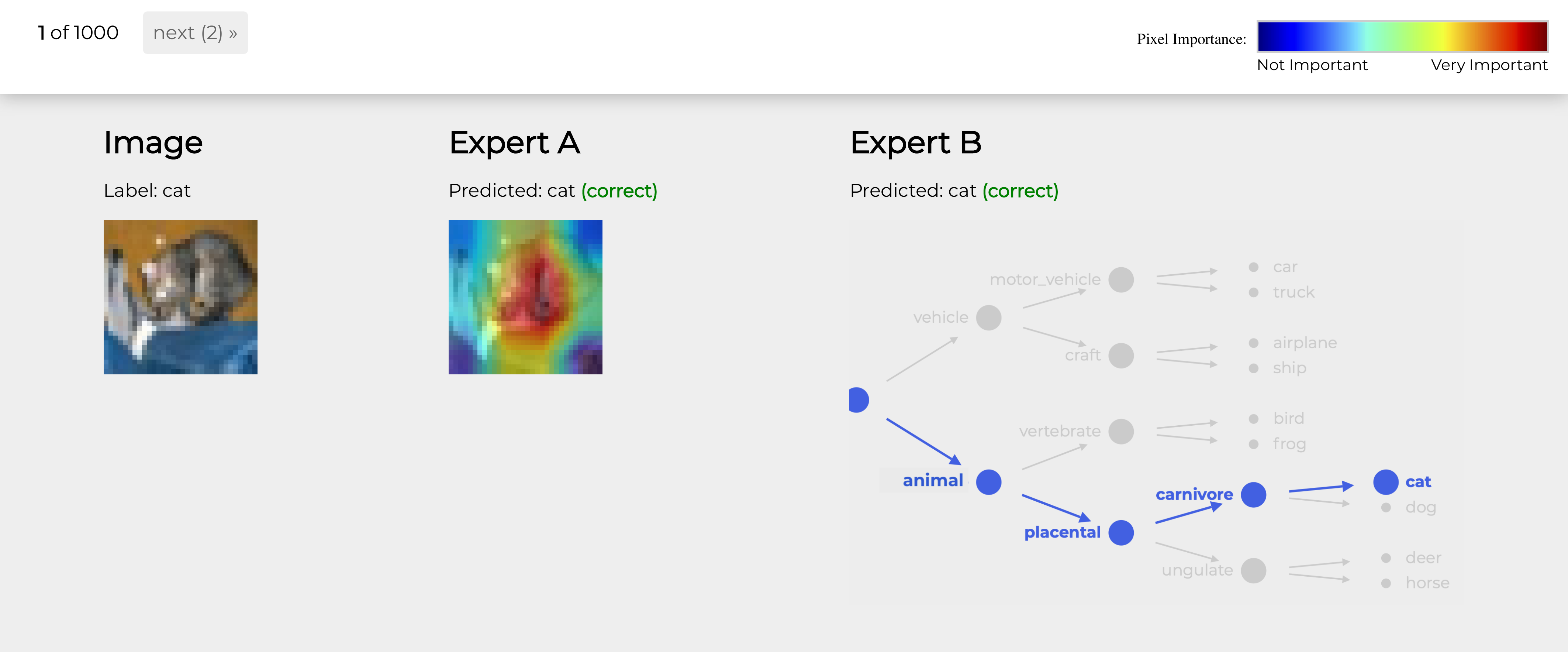}
    \caption{An example of a survey question presented to mechanical turks.}
    \label{fig:survey}
\end{figure}

\section{CIFAR100 Tree Visualization}
\begin{figure}[t]
    \centering
    \includegraphics[width=1.2\textwidth, trim={0cm 0cm 0cm 0cm}]{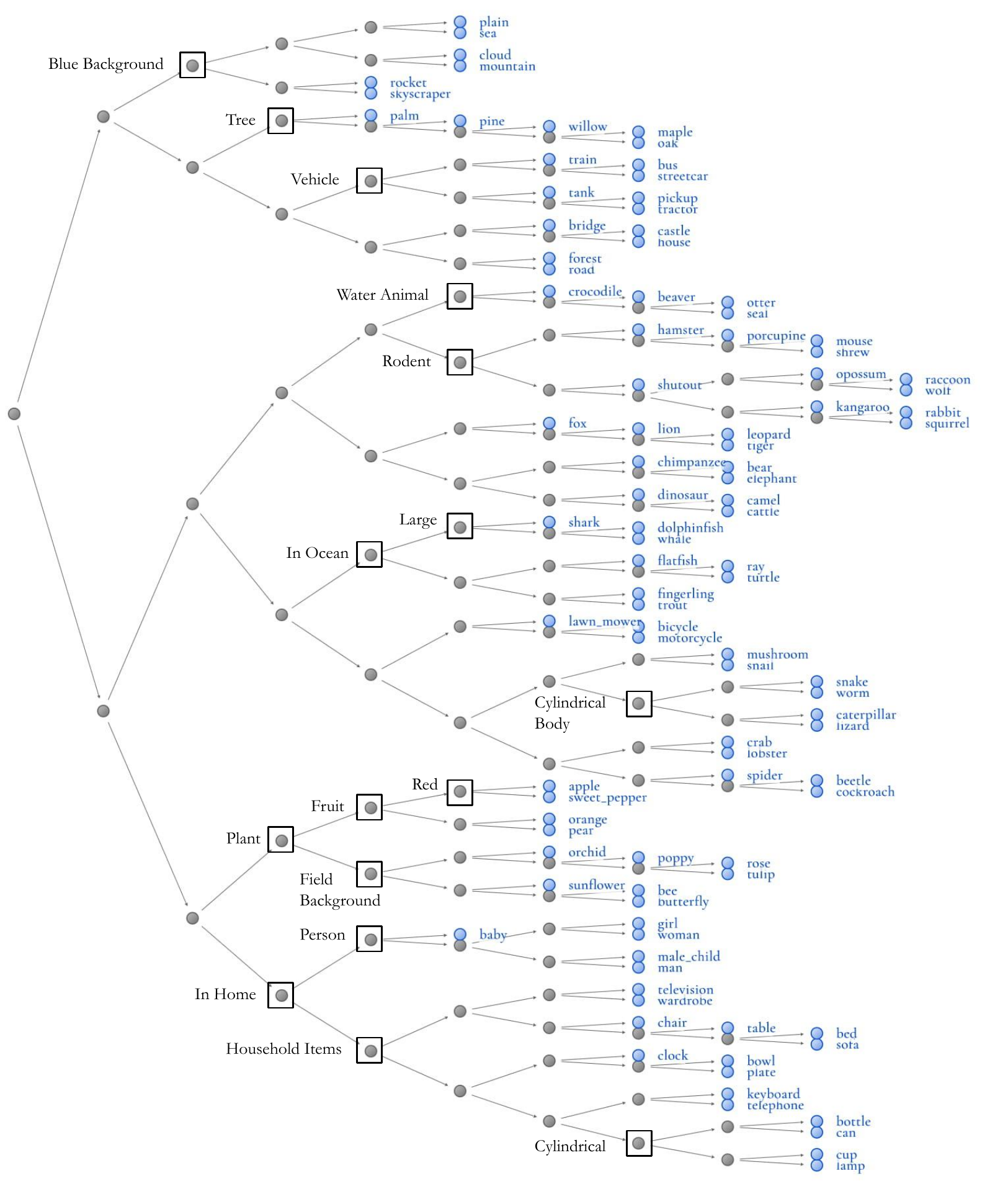}
    \caption{CIFAR100 tree visualization on WideResNet28x10 with samples of intermediate node hypothesis. Some nodes split on Wordnet attributes while other split on visual attributes like color, shape, and background.}
    \label{fig:cifar100_tree}
\end{figure}

We presented the tree visualizations for various models on the CIFAR10 dataset in Sec. 5 of the paper. Here we also show that similar visual meanings can be drawn from intermediate nodes of larger trees such as the one for CIFAR100. Figure~\ref{fig:cifar100_tree} displays the tree visualization for a WideResNet28x10 architecture on CIFAR100 (same model listed in Table 1 of Sec. 4.2). It can be seen in Figure~\ref{fig:cifar100_tree} that subtrees can be grouped by visual meaning, which can be a Wordnet attribute like \textit{Vehicle} or \textit{Household Item}, or a more contextual meaning such as shape or background like \textit{Cylindrical} or \textit{Blue Background}. 

\end{document}